\documentclass{article}
\pdfoutput=1
\usepackage{arxiv}

\usepackage[utf8]{inputenc} % allow utf-8 input
\usepackage[T1]{fontenc}    % use 8-bit T1 fonts
\usepackage{hyperref}       % hyperlinks
\usepackage{url}            % simple URL typesetting
\usepackage{booktabs}       % professional-quality tables
\usepackage{amsfonts}       % blackboard math symbols
\usepackage{nicefrac}       % compact symbols for 1/2, etc.
\usepackage{microtype}      % microtypography
\usepackage{lipsum}

\usepackage{amsmath}
\usepackage{graphicx,psfrag,epsf}
\usepackage{enumerate}
\usepackage{natbib}

\usepackage{adjustbox}
\usepackage{multirow}
\usepackage{xcolor}
\usepackage{caption}
\usepackage{subcaption}
\usepackage{chngcntr}
\usepackage[flushleft]{threeparttable}

\renewcommand{\theequation}{\arabic{equation}}

\title{\bf Deep Learning for Quantile Regression under Right Censoring: DeepQuantreg}

\author{
Yichen Jia\\
Department of Biostatistics\\
Graduate School of Public Health\\
University of Pittsburgh, Pittsburgh, USA\\
\And
Jong-Hyeon Jeong \thanks{Corresponding Author: jjeong@pitt.edu}\\
Department of Biostatistics\\
Graduate School of Public Health\\
University of Pittsburgh, Pittsburgh, USA\\
}

\begin{document}
	\maketitle

\begin{abstract}
The computational prediction algorithm of neural network, or deep learning, has drawn much attention recently in statistics as well as in image recognition and natural language processing. Particularly in statistical application for censored survival data, the loss function used for optimization has been mainly based on the partial likelihood from Cox's model and its variations to utilize existing neural network library such as Keras, which was built upon the open source library of TensorFlow. This paper presents a novel application of the neural network to the quantile regression for survival data with right censoring, which is adjusted by the inverse of the estimated censoring distribution in the check function. The main purpose of this work is to show that the deep learning method could be flexible enough to predict nonlinear patterns more accurately compared to existing quantile regression methods such as traditional linear quantile regression and nonparametric quantile regression with total variation regularization, emphasizing practicality of the method for censored survival data.
Simulation studies were performed to generate nonlinear censored survival data and compare the deep learning method with existing quantile regression methods in terms of prediction accuracy. The proposed method is illustrated with two publicly available breast cancer data sets with gene signatures. The method has been built into a package and is freely available at \url{https://github.com/yicjia/DeepQuantreg}.
\end{abstract}
%\vspace{.3cm}

\section{Introduction}
\label{s:intro}

Deep learning algorithms have been developed to perform classification and prediction in many application areas such as image recognition and natural language processing \citep{bishop1995neural,gallant1993neural}. Recently deep learning has gained much popularity in survival analysis because of its flexible model design and ability of capturing nonlinear relationships. The main challenge of applying deep neural network models to time-to-event data is the presence of censoring. To overcome this challenge, many deep learning models propose to use the Cox proportional hazards (PH) model-based, or the partial likelihood-based \citep{cox1975partial}, loss function for predicting patient survival \citep{faraggi1995neural,katzman2018deepsurv,ching2018cox}. \citet{faraggi1995neural} replaced the linear function in the partial likelihood with a nonlinear functional form of the output in the neural network algorithm, extending the Cox's model to a non-linear proportional hazards modeling. Alternatively, discrete-time survival models with neural network have been recently developed \citep{fotso2018deep, lee2018deephit, giunchiglia2018rnn}, which is more flexible than the Cox model-based methods and can predict the survival probability within the pre-specified time interval. However, the continuous nature of the data is lost in this way, since the survival time is discretized into finitely many pre-determined values. In addition to the restrictive assumption in the Cox PH model, the interpretation of the hazard function could be also delicate for non-statisticians like clinicians and patients. The accelerated failure time (AFT) model, which directly links the logarithm of event time to the covariates or predictors, have become a popular alternative possibly due to more straightforward interpretation. However, the AFT model is often modeled parametrically in that the logarithm of event time would be linearly linked with predictors with a specific error distribution such as a normal distribution, an extreme value distribution, or a logistic distribution, or semiparametrically modeled with the restrictive assumption that the baseline survival function is accelerated in time by the exponentiated covariate effects. Therefore, the quantile-based approach could practically have some advantages over the AFT and hazard-based models due to its nonparametric nature.

Without any covariates or predictors, for a positive random variable $T$ the quantile is generally defined as 
$$Q_T(\tau) = \text{inf}\{t: \mbox{Pr}(T \leq t ) \geq \tau \}, \tau \in (0,1).$$ In practice, however, investigators would be more interested in associations between the quantiles of time-to-event distributions and potential predictors in a regression setting. Quantile regression, originally proposed by \citet{koenker1978regression}, is a popular alternative to the least-square approach in the linear regression. It also has been an attractive alternative to the Cox PH model for time-to-event data, and numerous methods have been established to deal with the censoring problem under the quantile regression \citep{powell1984least,ying1995survival,mckeague2001median,peng2008survival}. Since the quantile is directly defined from the cumulative distribution function or the survival function, the whole spectrum of the quantiles or percentiles can be estimated given the covariate values, providing the entire shape of the survival distribution, while allowing for statistical inference on specific percentiles of interest, if necessary.

\citet{cannon2011quantile} has developed an \verb!R! package \verb!QRNN!, which implements the quantile regression neural network for continuous response variable, i.e. precipitation amounts truncated at zero, but it cannot be used to incorporate the random censorship commonly encountered in survival data. To our best knowledge, there is no literature on deep learning method for the quantile regression on right-censored survival data. In this paper, we present a novel deep censored quantile regression that is flexible to fit both log-linear and log-nonlinear time-to-event data for the purpose of more accurate prediction compared to existing quantile regression methods. The proposed method using the Huber check function with inverse probability weights has been implemented via Python library \verb!Keras!, which is the high-level Application Programming Interface (API) of \verb!TensorFlow 2.0!, and it is available on GitHub at \url{https://github.com/yicjia/DeepQuantreg}.

In Section 2, we review the existing work on censored quantile regression. In Section 3, we present the explanation and implementation of our deep censored quantile regression. The proposed algorithm is assessed via simulation studies in Section 4 and illustrated with two breast cancer data sets in Section 5. Finally, we conclude our paper with a brief discussion in Section 6.

\section{Censored Quantile Regression}

In this section, we review existing quantile regression methods for censored survival data. 

\subsection{Without censoring}

Suppose data consist of a positive continuous time to an event of interest $T$ and a covariate vector $\boldsymbol{x}^{'}=(1,x_1,...,x_p)$ associated with a regression coefficient parameter vector $\beta^{'}=(\beta_0,\beta_1,...,\beta_p)$. By definition, the $\tau^{th}$ conditional quantile function of the dependent variable $T$ given covariates $\boldsymbol{x}$ is defined as 
$$Q_{T|\boldsymbol{x}}(\tau) = \text{inf}\{t: \mbox{Pr}(T \leq t |\boldsymbol{x}) \geq \tau \}, \tau \in (0,1).$$
Suppose $(T_i,\boldsymbol{x_i})$ is a realization of the random variable $T$ without censoring and a covariate vector for the $i^{th}$ subject. Then the popular log-linear quantile regression model can be specified as
\begin{equation}
    Q_{T_i|\boldsymbol{x_i}}(\tau)= \exp(\beta_{\tau}^{'}\boldsymbol{x_i}). \label{eqn;2.1}
\end{equation}
The regression parameters in model (\ref{eqn;2.1}) are commonly estimated by minimizing the sum of the absolute deviations (LAD, least absolute deviation)
\begin{equation}
\sum_{i=1}^n |\log(T_i)-\beta^{'}_{\tau} \boldsymbol{x_i}|.\label{eqn;2.3}
\end{equation}
Equivalently the LAD estimators can be obtained by minimizing the check function \citep{koenker1978regression}
\begin{equation} 
\rho_{\tau}(u) = u[\tau - I(u \leq 0)], \label{eqn;2.4}
\end{equation}
where $u=\log(T_i) - \beta^{'}_{\tau}\boldsymbol{x_i}$. 

\subsection{Comparison with the accelerated failure time (AFT) model}
In this section, we point out a delicate difference between the quantile regression model (\ref{eqn;2.1}) and the AFT model. Let us denote $\boldsymbol{x_i}^{\prime}=(1,x_{i1},x_{i2},...,x_{ip})=(1,\boldsymbol{z_i}^{\prime})$, where $\boldsymbol{z_i}^{\prime}=(x_{i1},x_{i2},...,x_{ip})$, and $\boldsymbol{\beta}^{\prime}=(\beta_0,\beta_1,\beta_2,...,\beta_p)=(\beta_0,\boldsymbol{\gamma}^{\prime})$, where $\boldsymbol{\gamma}^{\prime}=(\beta_1,\beta_2,...,\beta_p)$. Then since $\log(T_i)=\beta_0+\boldsymbol{\gamma}^{\prime} \boldsymbol{z_i}+\epsilon_i$
implies $T_i=e^{\beta_0+\epsilon_i+\boldsymbol{\gamma}^{\prime} \boldsymbol{z_i}}$, the conditional survival function given the covariates is 
\begin{equation}
S_{T_i}(t|\boldsymbol{x_i})=\mbox{Pr}(T_i > t|\boldsymbol{x_i})
= \mbox{Pr}(e^{\beta_0+\epsilon_i} > t e^{-\boldsymbol{\gamma}^{\prime} \boldsymbol{z_i}} |\boldsymbol{x_i})=S_{T0}(t e^{-\boldsymbol{\gamma}^{\prime}\boldsymbol{z_i}}), \label{eqn;2.21}
\end{equation}
where $S_{T0}(\cdot)$ is the baseline survival function of $e^{\beta_0+\epsilon}$, i.e. when $\boldsymbol{\gamma}^{\prime} \boldsymbol{z_i}=0$. Note that the model (\ref{eqn;2.21}) specifies the AFT model after the intercept term $\beta_0$ has been absorbed into the baseline survival function or an unrealistic assumption of no intercept. Therefore, under the AFT model, it might be difficult to estimate the intercept nonparametrically, and it seems crucial to estimate the baseline survival function correctly because the remaining estimates of the regression parameters could be biased otherwise. 

The AFT model in (\ref{eqn;2.21}) induces a proportional quantile regression model
\begin{equation}
Q_{T_i|\boldsymbol{z_i}}(\tau) = Q_{T0}(\tau) \exp{(\boldsymbol{\gamma}_{\tau}^{'}\boldsymbol{z_i})}, \label{eqn;2.2}
\end{equation}
because $S_{T_i}(t|\boldsymbol{z_i})$ maps $Q_{T_i|\boldsymbol{z_i}}(\tau)$ to $1-\tau$ and the baseline quantile function can be expressed as $Q_{T0}(\tau)=S^{-1}_{T0}(1-\tau)$. For the simplest case with a single binary covariate as a group indicator $x_1=0$ (control) or 1 (intervention), the model (\ref{eqn;2.2}) can be specified as
$Q_{T_i|x_1=0}(\tau)=Q_{T0}(\tau)$ and $Q_{T_i|x_{1}=1}(\tau)=Q_{T0}(\tau) \exp{(\beta_1)}$, respectively, so that $\beta_{1}$ can be interpreted as the log-ratio or the log-difference of the two quantile functions regardless of the baseline quantile function $Q_{T0}(\tau)$. The same interpretation can come from the model (\ref{eqn;2.1}) after the intercept term being canceled out, but a delicate difference between the two models arises in terms of explicit presence of the intercept in the model. It should be cautious that the common acceptance of equivalence of the log-linear model to the AFT model holds only under the assumption that the intercept term from the log-linear model has been integrated into the baseline survival distribution in the AFT model and it needs to be implicitly yet consistently estimated through the baseline survival function. Therefore, in general using the model (\ref{eqn;2.1}) would be more advantageous, straightforward, and efficient in estimating all the covariate effects including the intercept on the quantiles.

\subsection{Under right censoring}
\label{s:rightcensoring}
We now consider right-censored survival data. Let $T_i$ and $C_i$ denote potential failure time and potential censoring time, respectively, and they are independent conditional on the covariate vector $\boldsymbol{x_i}$. In many clinical trials and biomedical studies, we only observe ($Y_i$, $\delta_i$, $\boldsymbol{x_i}$), where $Y_i = \text{min}(T_i,C_i)$ is the observed survival time, and $\delta_i = I(T_i \leq C_i)$ is the event indicator. To incorporate the right-censoring, assuming the conditional independence between $T_i$ and $C_i$ given $\boldsymbol{x_i}$ and the independence of $C_i$ from $\boldsymbol{x_i}$, we include a weight function 
$$\omega_i = \frac{\delta_i}{\hat{G}(Y_i)},$$
in (\ref{eqn;2.4}) where $\hat{G}(\cdot)$ is the Kaplan-Meier estimator of the censoring distribution based on the observed data $\{Y_i, I(\delta_i=0)\}$. This weight function can be shown to be equivalent to jumps of the Kaplan-Meier estimator of the distribution function of the logarithm of the failure time \citep{huang2007least}. Therefore, the estimator $\hat{\beta}_{\tau}$ is the minimizer of  
\begin{equation}
    \sum_{i=1}^n \omega_i\cdot\rho_\tau(\log(Y_i) - \beta^{'}_{\tau}\boldsymbol{x_i}).
\label{eqn:cquantreg}
\end{equation}
The solutions to equation (\ref{eqn:cquantreg}) can be obtained by the function \verb!rq.wfit()! with the option of specifying the weights $\omega_i$ from the \verb!quantreg! package in R. The consistency and asymptotic normality of the estimator $\hat{\beta}_{\tau}$ has been established in \citet{huang2007least}. 

The assumption of independence between the censoring distribution and covariates may seem strong, but it often holds for data from well-conducted clinical trials with administrative censoring. When the assumption is not met, i.e. the censoring distribution and covariates are dependent, the conditional survival function of $G(\cdot|\boldsymbol{x_i})$, can be estimated from a stratified or adjusted Kaplan-Meier estimator \citep{xie2005adjusted} or imposing the Cox PH model for $C_i$ given $\boldsymbol{x_i}$.

\subsection{Nonparametric Quantile Regression-Review}
When the relationship between the response variable and covariates are non-linear, it is common to infer the appropriate functional form using a data-driven approach, and estimate the effects of covariates on the response variable nonparametrically. Smoothing spline is one of the most popular and powerful techniques in nonparametric regression \citep{eubank1988spline}. In the quantile regression, \citet{koenker1994quantile} considered the general form of the additive models
$$Q_{Y_i|x_i,z_i}(\tau|x_i,z_i) = \beta^{'}_{\tau}x_i + \sum_{j=1}^J g_j(z_{ij}),$$
where  $g_j(z_{ij})$'s ($j=1,2,...,J$) are assumed to be the continuous functions for the $i^{th}$ covariate $z_i$ that needs to be nonparametrically estimated. The objective function with total variation regularization \citep{rudin1992nonlinear} to be minimized for each quantile smoothing spline term is
$$\sum_{i=1}^n \rho_\tau(y_i - g(z_{i})) + \lambda V(g'(z)),$$
where $V(g'(z))=\int_0^1 |g^{\prime \prime}(z)|dz$ denotes the total variation of the derivative of the function $g$ for the univariate case. The smoothing parameter $\lambda$ controls the trade-off between fidelity and the penalty component, where a larger $\lambda$ leads to a smoother fit. This procedure has been implemented as the \texttt{rqss()} function in \texttt{quantreg} package in \texttt{R}. The \texttt{rqss()} allows additive nonparametric terms (quantile smoothing spline terms) in the quantile regression through the function \texttt{qss()}. Qualitative constraints can be also specified in the fitted \texttt{qss()} function, e.g. increasing, decreasing, convex or concave, restricting the coefficients of the linear basis accordingly. Even though the nonparametric quantile regression can fit nonlinear data, it seems that the shape of the nonlinearty needs to be restricted by the qualitative constraints. If no constraint is specified, the fitted curve may be over-smoothed, which could lead to poor prediction result.

\section{Deep Censored Qunatile Regression: DeepQuantreg}
\subsection{Model Architecture}
\begin{figure}
    \centering
    \includegraphics[width = 0.9\linewidth,keepaspectratio]{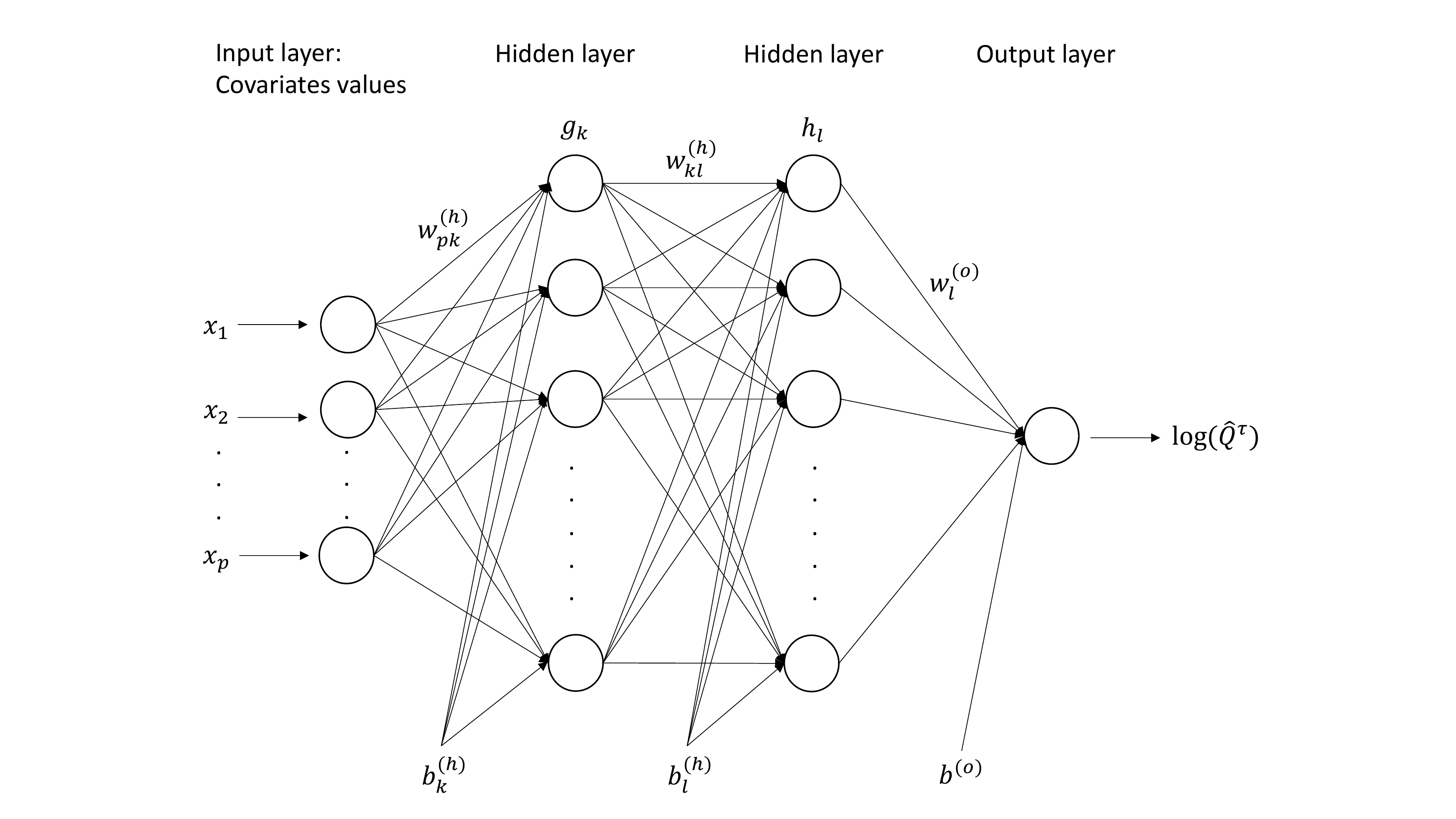}
    \caption{An overview of the deep censored quantile regression architecture used in this study.}
    \label{fig:architecture}
\end{figure}

We propose a deep feed-forward neural network to predict the conditional quantile. Figure \ref{fig:architecture}  shows the basic model architecture. The input to the network is the covariate vector $x_{p}$ ($p=1,2,...,P$). The hidden layers of the network are dense, i.e. fully connected by the nodes. The output of the hidden layer is given by applying the activation function to the inner product between the input and the hidden-layer weights plus the hidden-layer bias. For example, suppose there are $P$ input variables and two hidden layers. Then, the output of the $k^{th}$ hidden node for the first hidden layer would  be 
$$g_k = f_1\left(\sum_{p=1}^{P} x_{p} w_{pk}^{(h)}+b_k^{(h)}\right),\quad k=1,2,...,K,$$
and the output of the $l^{th}$ hidden node in the second hidden layer would be 
$$h_l = f_2\left(\sum_{k=1}^K g_k w_{kl}^{(h)}+b_l^{(h)}\right),\quad l=1,2,...,L,$$ 
where $f_1(\cdot)$ and $f_2(\cdot)$ denotes the activation functions for hidden layers, and $w^{(h)}$ and $b^{(h)}$ represent the hidden-layer weights and bias, respectively, both of which get updated at each training iteration. The bias terms allow for shifting the activation function outputs left or right \citep{gallant1993neural, bishop1995neural,reed1999neural}.
Lastly, the output layer of the network is a single node with a linear activation function which gives the estimate of the conditional $\tau^{th}$ quantile for the $i^{th}$ subject as
$$\log{\hat{Q}^{(\tau)}_i} = \sum_{l=1}^L h_{l,i} w_{l}^{(o)}+b^{(o)},$$
where $w^{(o)}$ and $b^{(o)}$ denote the output-layer weights and bias, respectively. Popular activation functions are the logistic function (sigmoid), rectified linear unit (RELU) defined as $\max(0,y)$, where $y$ is a linear response function, and scaled exponential linear unit (SELU), among other things. It is known that the sigmoid activation function can cause vanishing or exploding gradient problem, so the RELU or SELU is generally preferred.

\subsection{Cost, or Loss, Function}
Similar to the ordinary censored quantile regression, a modified form of the check function needs to be used in the cost function for the deep quantile regression. Following \citet{faraggi1995neural}, we replace the linear functional $\beta^{'}_{\tau}x_i$ in equation (\ref{eqn:cquantreg}) by the output from the neural network $\log{\hat{Q}^{(\tau)}_i}$, i.e.
\begin{equation}
    \sum_{i=1}^n \omega_i\cdot\rho_\tau(\log(Y_i) - \log{(\hat{Q}^{(\tau)}_i)}).
\label{eqn:DLloss}
\end{equation}
Note that, a single layer neural network with linear activation function could approximate the ordinary quantile regression.

\subsection{Optimization}
The optimization of the neural network is achieved by the gradient descent process in the backpropagation phase \citep{rumelhart1986learning} to minimize the cost, or loss, function. The minimum can be obtained as the derivative of the loss function with respect to the weights through the chain rule involving the inputs and outputs of the activation functions reaches 0. Since it is often impossible to obtain a closed form solution for the weights including the biases in the deep learning setting, however, the gradient descent method is often used to iteratively search for the minimum of the loss function by updating old weight values by the amount of the slope of the loss function at those values multiplied by a learning parameter (usually small) that controls the size of convergence steps. The backpropagation algorithm through the gradient descent method was shown to be simple and computationally efficient \citep{Goodfellow-et-al-2016}. Optimizers implemented in TensorFlow typically include Stochastic Gradient Descent (SGD), SGD with Momentum, Adaptive gradient optimizer (AdaGrad) \citep{duchi2011adaptive}, AdaDelta \citep{zeiler2012adadelta}, Adam (adaptive moment estimation) \citep{kingma2014adam}, and Adamax (a variant of Adam based on the infinity norm), among other things. The initial values of the weights for these algorithms could be generated from some uniform or normal distributions that depend on the numbers of input and/or output variables, but more research seems to be needed.

The check function in equation (\ref{eqn;2.4}) is undefined at the origin, and thus not differentiable everywhere. Therefore, we adopted the Huber function \citep{huber1973robust} to smooth the check function as in \citet{chen2007finite} and \citet{cannon2011quantile}, which is defined as 
\begin{equation}
    \rho_{\tau}(u) = 
    \begin{cases}
      \tau h(u) \quad \text{if} \quad u \geq 0\\
      (\tau-1) h(u) \quad \text{if} \quad u < 0 \\
    \end{cases} 
\label{eqn:hubercheck}
\end{equation}
where 
\begin{equation*}
    h(u) = 
    \begin{cases}
      \frac{u^2}{\xi} \quad \text{if} \quad 0 \leq |u| \leq \xi\\
      |u| - \frac{\xi}{2} \quad \text{if} \quad |u| > \xi \\
    \end{cases} 
\end{equation*}
is the Huber function.

\subsection{Hyperparameter Tuning}
Hyperparameter tuning is essential in machine learning methods such as neural network and random forests since each different training run of the algorithm could provide a different output even with the same set of hyperparameters. As will be shown later, in addition, a model with more layers and more nodes per layer tends to capture the nonlinear patterns in the data better, so hyperparameter tuning is also crucial to prevent overfitting. In this paper, hyperparameters involved in neural network, including number of hidden layers (1, 2, and 3 layers), number of nodes in each layer (100, 300, and 500 nodes), activation function (RELU, Sigmoid, Hard Sigmoid, and Tanh), optimizer (Adam, Nadam, Adadelta, and AdaMax), dropout rate (0.1, 0.2, 0.3, and 0.4), number of epochs (200, 500, 800, and 1000) and batch size (32, 64, and 128), are tuned using 5-fold cross validation in the training data sets.

\subsection{Model Evaluation}
\label{metric}

The most common model evaluation methods compare the predicted and observed outcome values such as the mean squared error (MSE) without censoring, unless the true outcome values are known as in the simulation settings. To evaluate the prediction performance in this paper, we use the concordance-index ($C$-index) \citep{harrell1984regression}, modified mean squared error (MMSE), which is newly proposed here for the censored quantile regression, and modified check function \citep{li2017assessing}.

The $C$-index is one of the most common metrics used to assess the prediction accuracy of a model in survival analysis, which is a generalization of the area under the ROC curve (AUC) that takes censoring into account \citep{heagerty2005survival}. The $C$-index is also related to the rank correlation between the observed and predicted outcomes. Specifically, it is the proportion of all comparable pairs that the predictions are concordant. For example, two samples $i$ and $j$ are comparable if there is an ordering between the two possibly censored outcomes as $Y_i<Y_j \text{ and } \delta_i=1$. Then a comparable pair is concordant if a subject who fails at an earlier time point is predicted with a worse outcome, which is the predicted conditional quantiles in our case, i.e. $\hat{Q}^{(\tau)}_i < \hat{Q}^{(\tau)}_j$. The value of $C$-index is between 0 and 1 where 0.5 indicates a random prediction and 1 is a perfect association between predicted and observed outcomes. 

It is worth noting that the rank-based methods like the $C$-index are not sensitive to small differences in discriminating between two models \citep{harrell1996multivariable}. For instance, the $C$-index considers the (prediction, outcome) pairs (0.01, 0), (0.9, 1) as no more concordant than the pairs (0.05, 0), (0.8, 1). In other words, the $C$-index focuses on the order of the predictions instead of the actual deviations of the prediction from the observed outcome such as the MSE.

Similar to the MSE from the ordinary least squares method, the MMSE is defined as the mean of residual sum of squares between observed true event times and predicted quantile estimates under censoring. Since the true event times are not observed for the censored observations, the MMSEs are only calculated over event times. Note that by only considering the non-censored times, this metric would certainly benefit models that do not account for censoring, which is not an issue in this paper since all the methods we compared accounted for censoring by including the same inverse probability censoring weights (IPCW). Mathematically, for the traditional quantile regression, the MMSE can be defined as 
$$\sum_{i=1}^n \left[\delta_i \{\log(Y_i)-\hat{\beta}^{'}_{\tau} x_i\}\right]^2,$$
and for the neural network algorithm,
$$\sum_{i=1}^n \left[\delta_i \{\log(Y_i)-\log{(\hat{Q}^{(\tau)}_i)}\}\right]^2.$$

The expected check function, as in \citet{li2017assessing},
$$L(\tau) = E\rho_\tau(Y - \hat{Q}^\tau), $$
is another sensible measure for evaluating the difference between predicted and true quantiles. The comparison of $L(\tau)$ for multiple working models can reveal their relative prediction loss at the $\tau^{th}$ quantile. In our case, the plug-in estimator of $L(\tau)$ for the traditional quantile regression is 
$$\hat{L}_n(\tau) = \frac{1}{n}\sum_{i=1}^n \omega_i \cdot \rho_\tau(\log(Y^u_i) - \hat{\beta}^{'}_{\tau} x_i),$$
and for the neural network algorithm,
$$\hat{L}_n(\tau) = \frac{1}{n}\sum_{i=1}^n \omega_i \cdot \rho_\tau(\log(Y^u_i) - \log(\hat{Q}^{(\tau)}_i)),$$
where $Y^u_i = Y_i \wedge u$, $u$ being a pre-specified constant slightly smaller than the planned follow-up time, and $\omega_i$ is the IPCW defiend in Section \ref{s:rightcensoring}. The distributional properties and the corresponding inference procedures for $\hat{L}(\tau)$ has been derived in \citet{li2017assessing}.
It is worth mentioning that we use the observed time and covariates of a new subject from the same study population here, which distinguishes between using the check function as a model goodness-of-it criterion using the existing data and a model evaluation metric using new data points. The estimated expected check function will be referred to as ``quantile loss (QL)".

\subsection{Prediction Uncertainty}
\label{pi}
Understanding the uncertainty of a predicted outcome is also crucial in practice. It is worth noting that prediction interval is different from confidence interval in that the latter quantifies the uncertainty in an estimated population parameter while the former measures the uncertainty in a single predicted outcome. Several methods on obtaining the prediction interval have been proposed for the deep learning algorithm \citep{de1998prediction, nix1994estimating, heskes1997practical}. In this paper, we utilize dropout as Bayesian approximation of the Gaussian process to obtain model uncertainty \citep{gal2016dropout}. Dropout was first proposed as a regularization method to avoid over-fitting in neural network by randomly ``dropping out" a portion of nodes output in a given layer \citep{hinton2012improving, srivastava2014dropout}. Typically, dropout is only used during the training stage but not for the prediction with the fitted network. However, using Monte Carlo Dropout \citep{gal2016dropout}, i.e. enabling dropout at test time and repeating the prediction several times, provides the model uncertainty. This approach is suitable for any models with minimal changes that does not sacrifice computational complexity or test accuracy.

\section{Simulation Studies}
\label{s:sim}
In this section, we compare our deep censored quantile regression (DeepQuantreg) with the traditional censored quantile regression (Quantreg) and nonparametric quantile regression with total variation regularization through \texttt{rqss()} procedure, which  will  be  referred  to  as  the  rqss  method  throughout the paper, under different simulation scenarios. For the \texttt{rqss()} procedure, we specified a concave qualitative constraint, which best fits our simulated data, and we set $\lambda = 1$ which is the default setting. We also compare the predicted results with and without Huber function in our deep neural network model. Thus, the four models we compare are (i) traditional censored quantile regression with ordinary check function, (ii) nonparametric censored quantile regression, (iii) deep censored quantile regression with ordinary check function, and (iv) deep censored quantile regression with Huber check function. The hyperparameters we used in this simulation study are summarized in Supplementary Table S1.

\subsection{Data Generation Mechanism}

\begin{figure}
\begin{subfigure}{0.5\textwidth}
  %\centering
  % include first image
  \includegraphics[width=.95\linewidth]{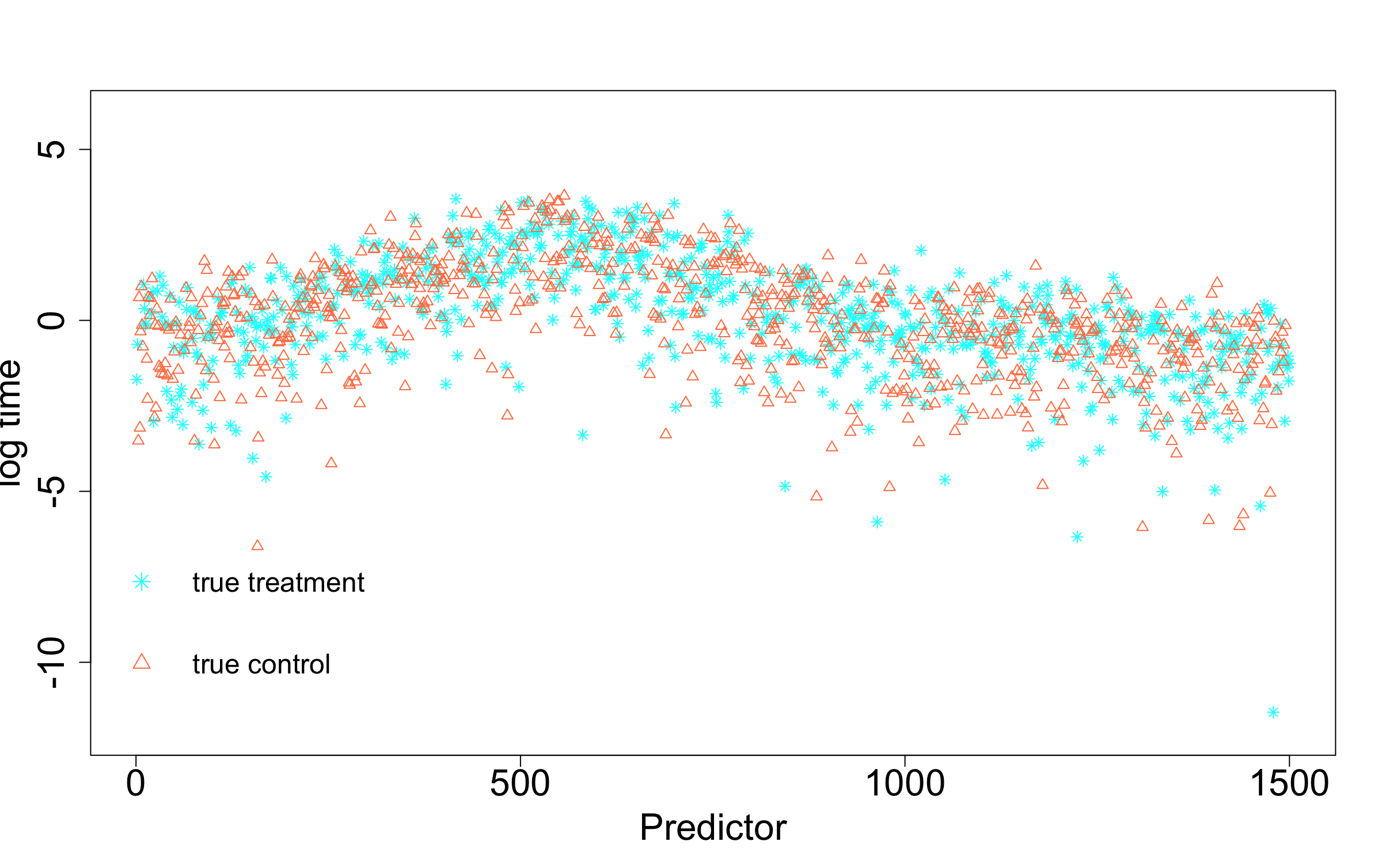}
  \caption{No group effect data with 10\% censoring}
  \label{fig:noeffect_rawdata10}
\end{subfigure}
\begin{subfigure}{0.5\textwidth}
  %\centering
  % include second image
  \includegraphics[width=.95\linewidth]{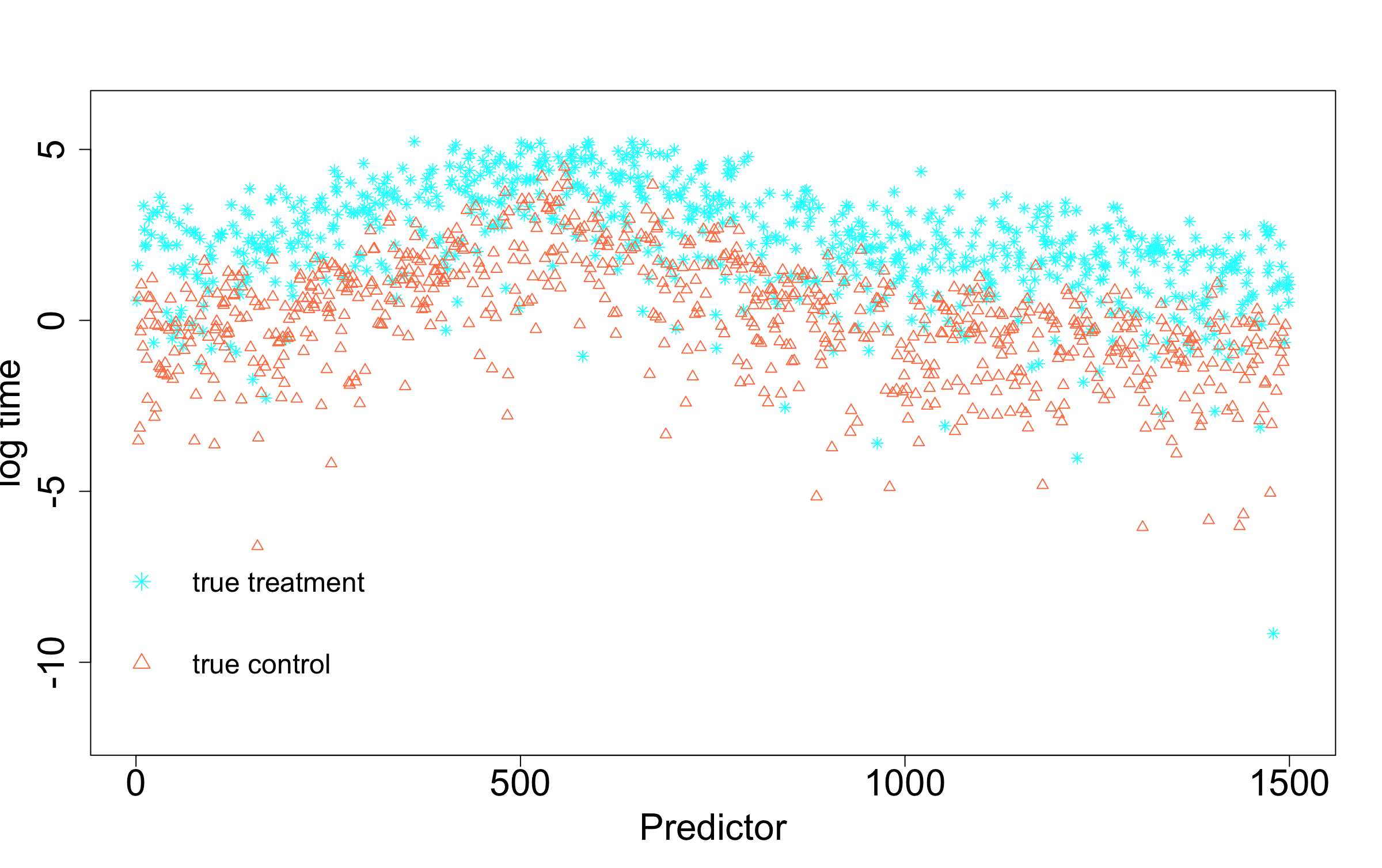}  
  \caption{Group effect data with 10\% censoring}
  \label{fig:groupeffect_rawdata10}
\end{subfigure}

\begin{subfigure}{0.5\textwidth}
  %\centering
  % include third image
  \includegraphics[width=.95\linewidth]{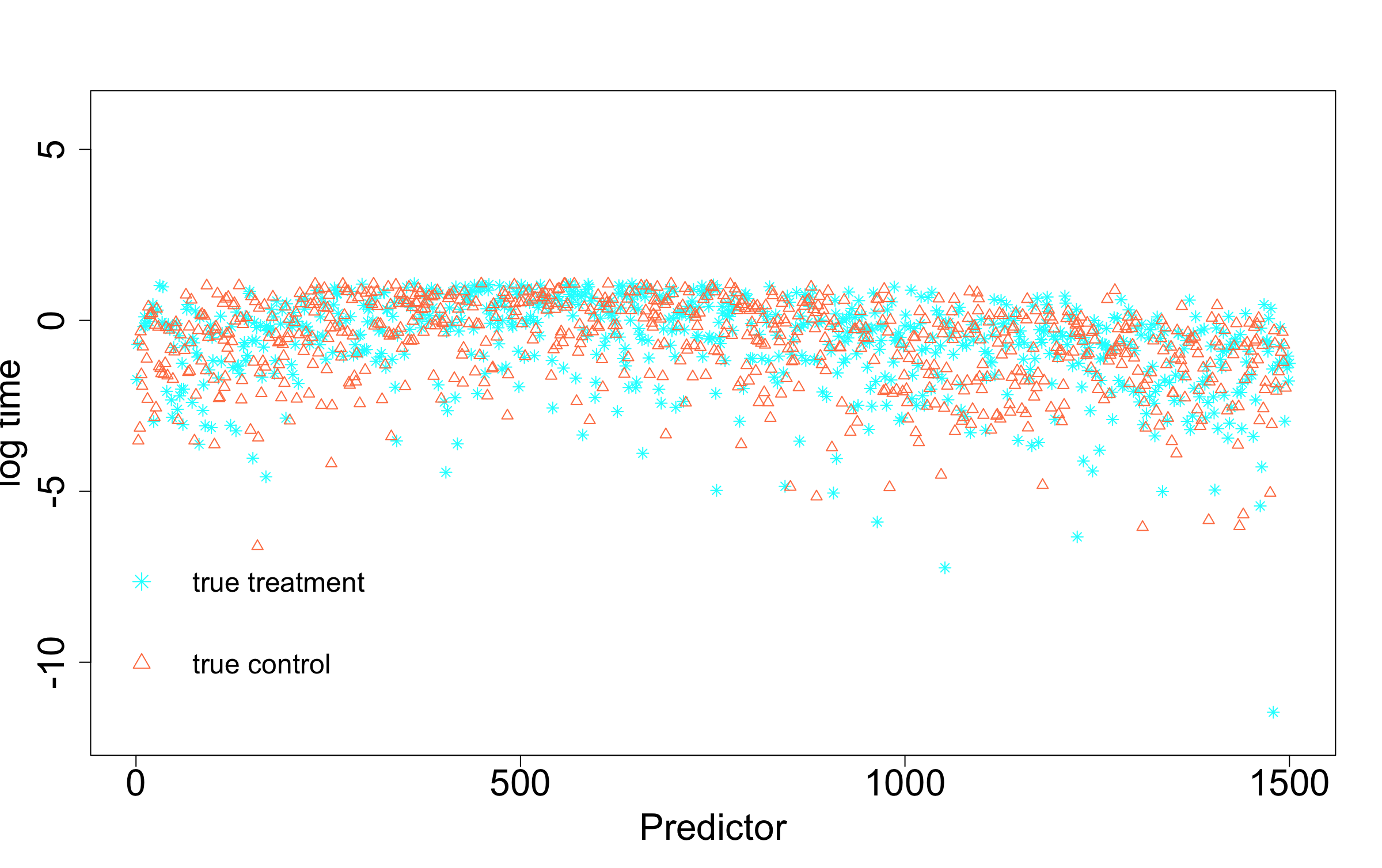}  
  \caption{No group effect data with 50\% censoring}
  \label{fig:noeffect_rawdata50}
\end{subfigure}
\begin{subfigure}{0.5\textwidth}
  %\centering
  % include third image
  \includegraphics[width=.95\linewidth]{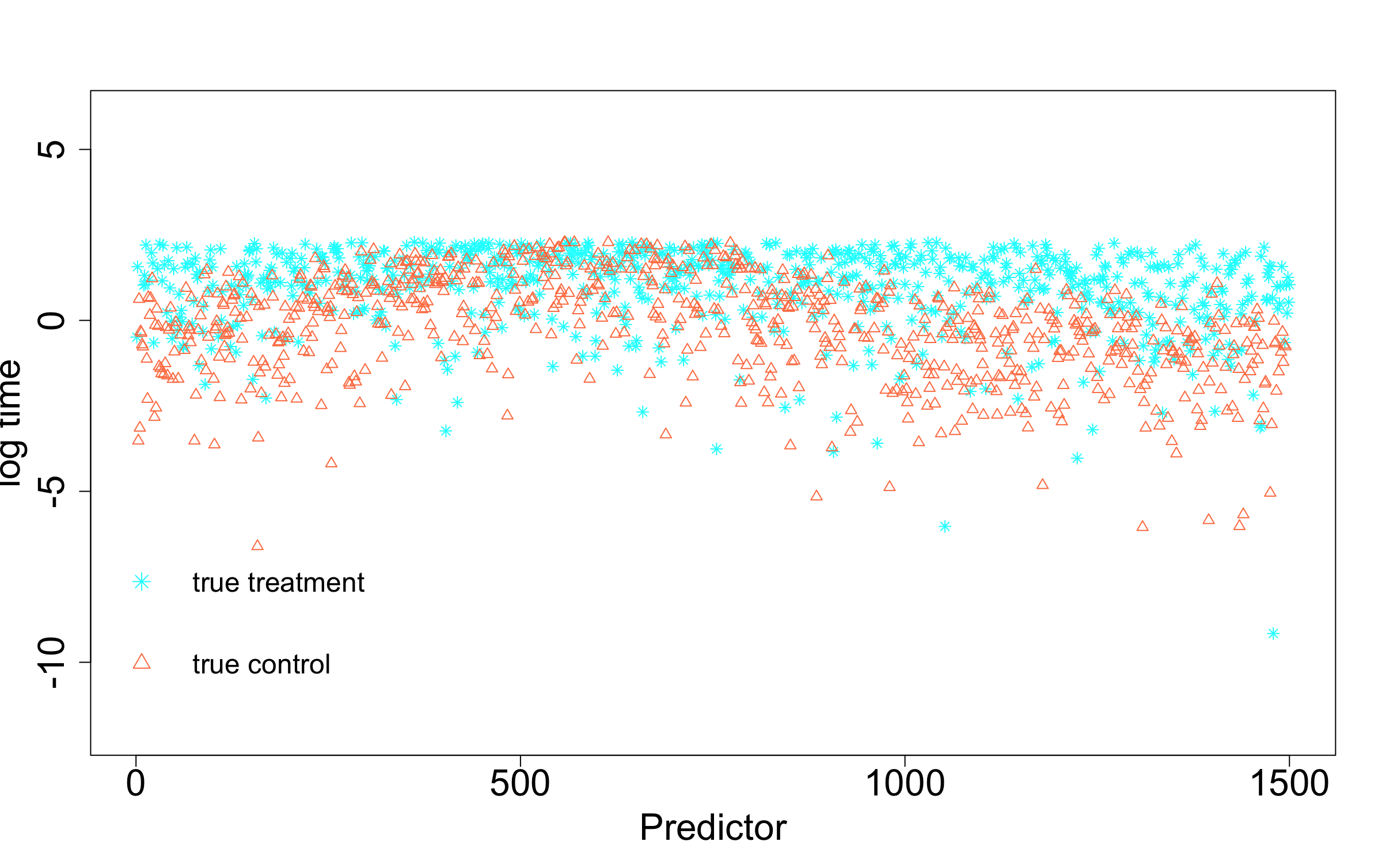}  
  \caption{Group effect data with 50\% censoring}
  \label{fig:groupeffect_rawdata50}
\end{subfigure}

\caption{Simulated data for no group effect data (left) and group effect data (right) under different censoring proportions}
\label{fig:rawdata}
\end{figure}

True failure times were first generated from piece-wise exponential distributions changing the event rates sequentially to induce nonlinear patterns without a group effect, which will be referred to as ``no group effect data". Second, a binary predictor was included to induce a multiplicative group effect, with the true failure times for the treatment group being a multiple of the true failure times for the control group, referred to as ``group effect data".  Finally, true censoring times were drawn from a uniform distribution between 0 and $c$, where $c$ controls the desired censoring proportions as 10\%, 30\%, 50\% or 70\%. The minimum of the true failure times and true censoring times were created as observed survival times, together with associated event indicators. For example, Figure \ref{fig:rawdata} shows specific realizations of simulated no group effect data and group effect data under 10\% (Figure \ref{fig:noeffect_rawdata10} and \ref{fig:groupeffect_rawdata10}) and 50\% censoring (Figure \ref{fig:noeffect_rawdata50} and \ref{fig:groupeffect_rawdata50}), respectively, where $\triangle$ indicates control group and $*$ does intervention group. Note that under 50\% censoring the larger true failure times tend to be censored in the middle. We generated training and test data sets with sample size of $n = 150, 750, \text{ and } 1500$ for both no group effect data and group effect data. We performed 1000 simulations and compared the $C$-index, the MMSE, and quantile loss metrics on the test sets.

\subsection{Main Comparison}
\label{maincomparison}
\begin{figure}
\begin{subfigure}{0.5\textwidth}
  %\centering
  % include first image
  \includegraphics[width=0.95\linewidth]{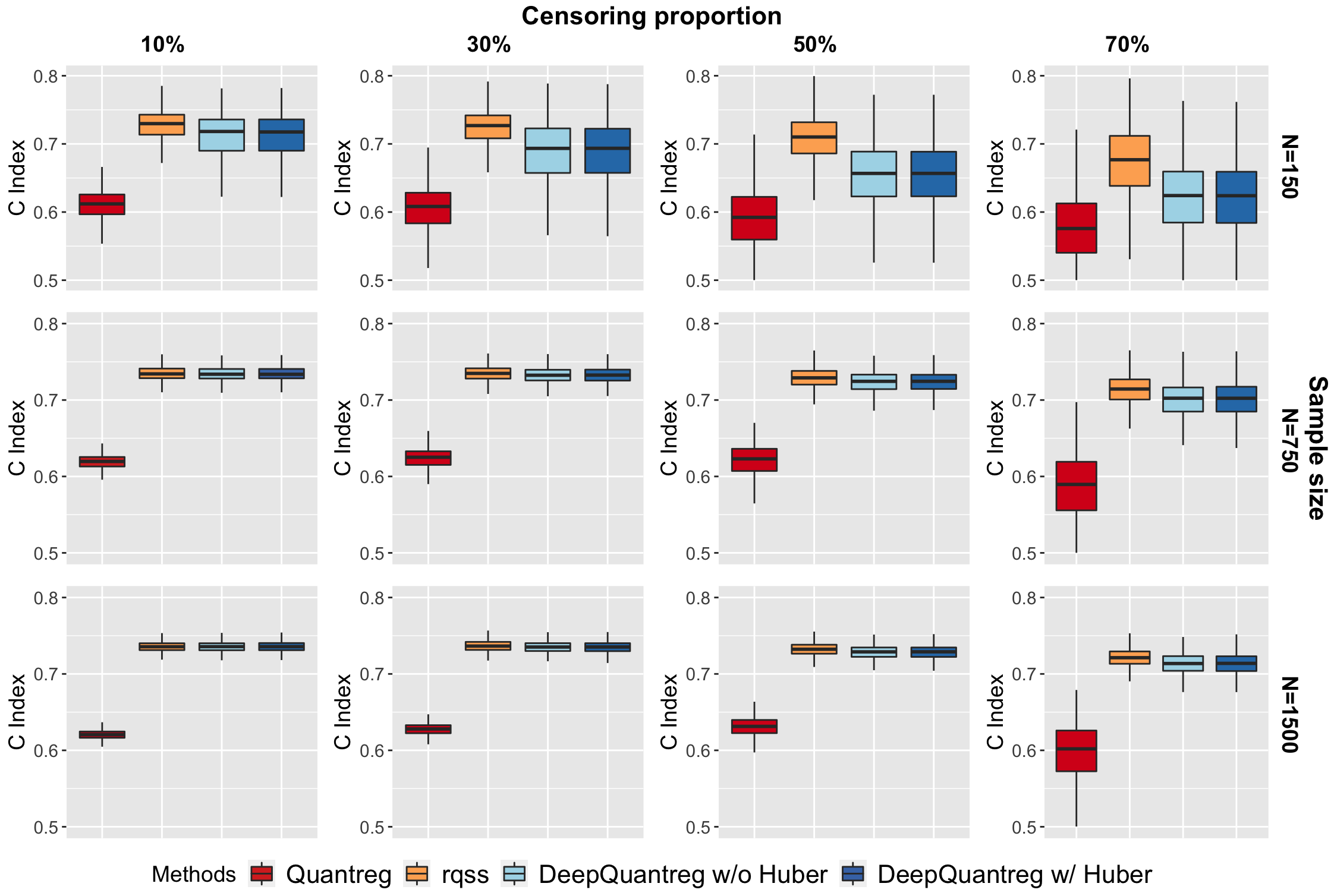}
  \caption{$C$-Index for no group effect data}
  \label{fig:noeffect_cindex}
\end{subfigure}
\begin{subfigure}{0.5\textwidth}
  %\centering
  % include first image
  \includegraphics[width=0.95\linewidth]{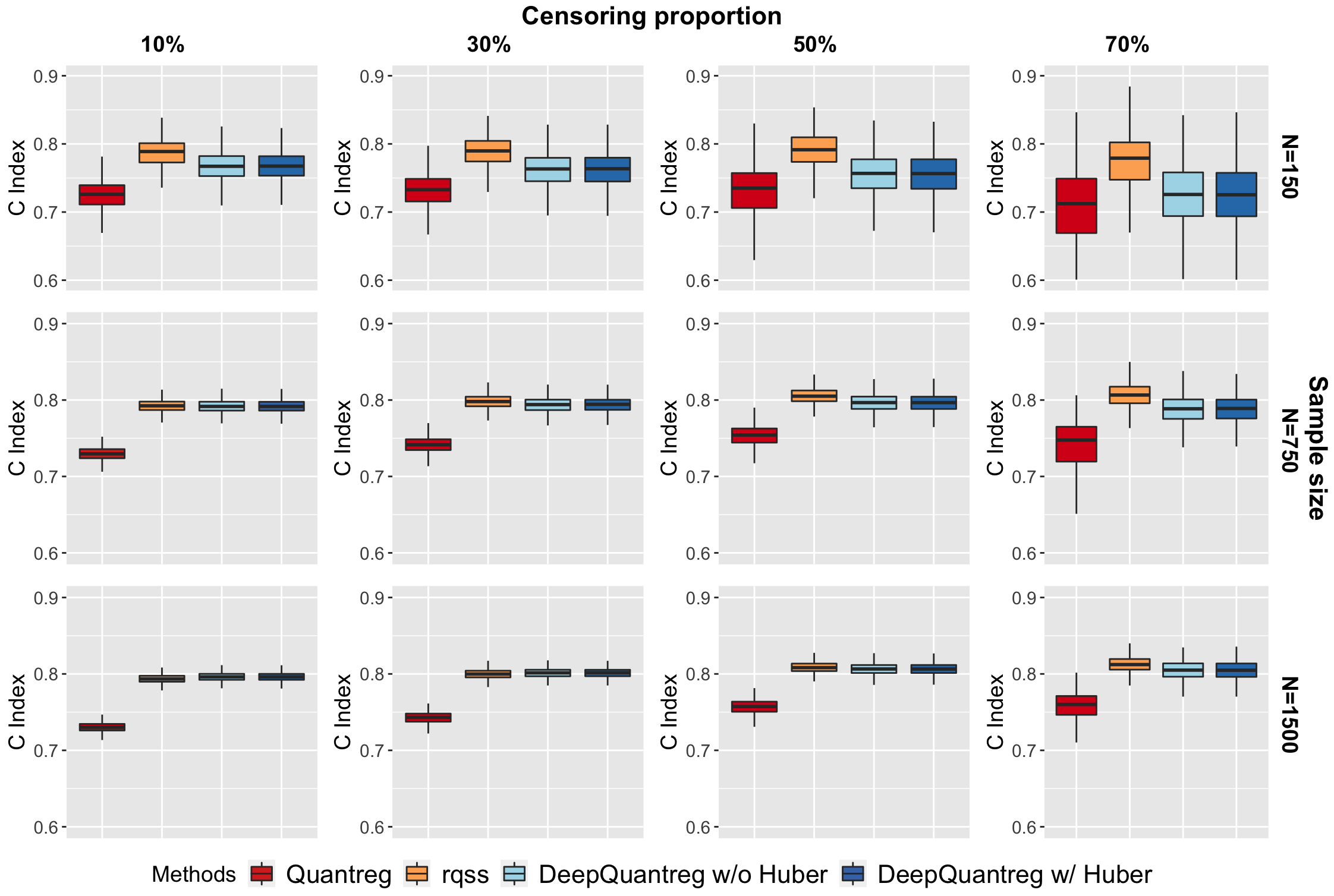}
  \caption{$C$-Index for group effect data}
  \label{fig:groupeffect_cindex}
\end{subfigure}

\begin{subfigure}{0.5\textwidth}
  %\centering
  % include second image
  \includegraphics[width=.95\linewidth]{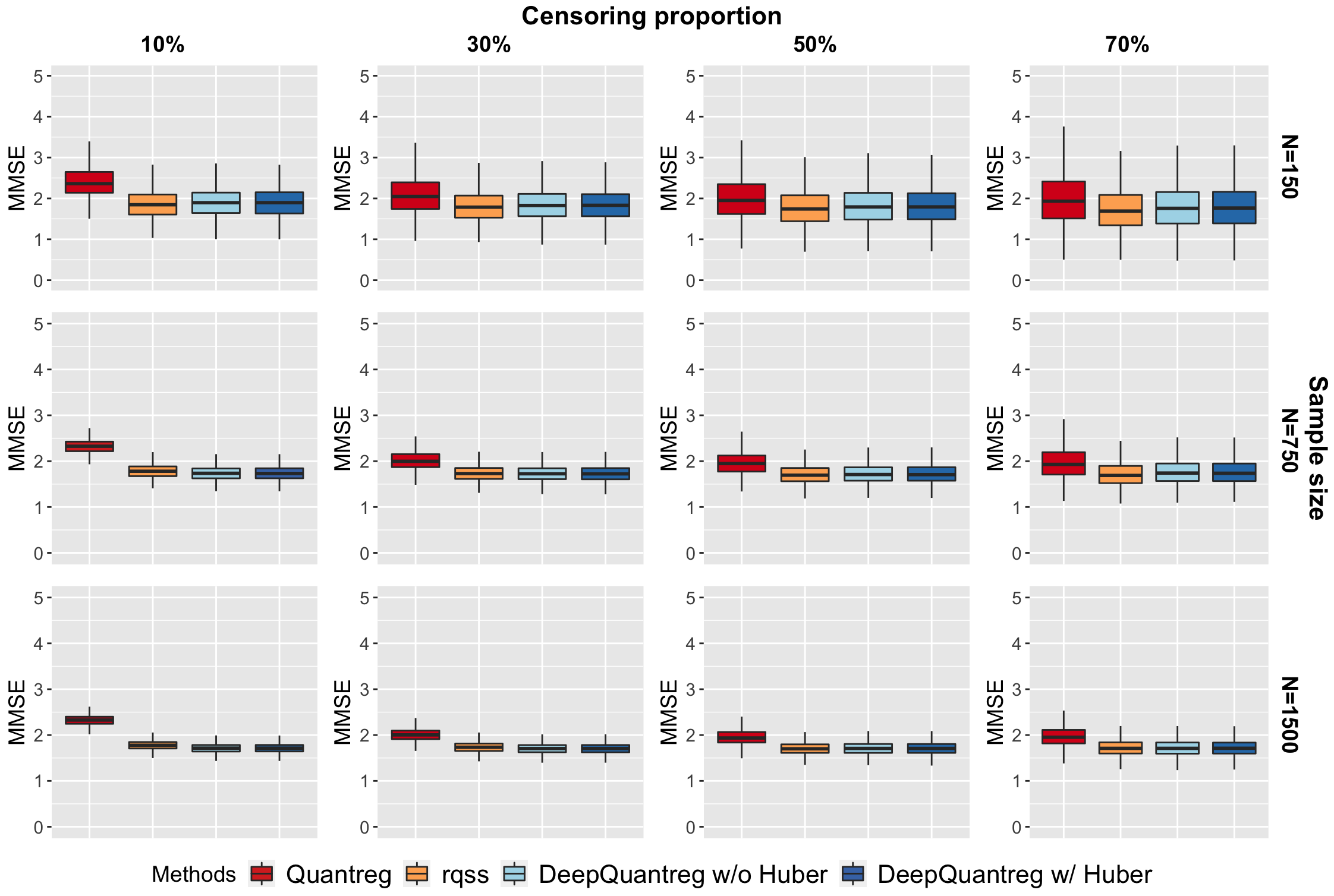}  
  \caption{MMSE for no group effect data}
  \label{fig:noeffect_mmse}
\end{subfigure}
\begin{subfigure}{0.5\textwidth}
  %\centering
  % include first image
  \includegraphics[width=.95\linewidth]{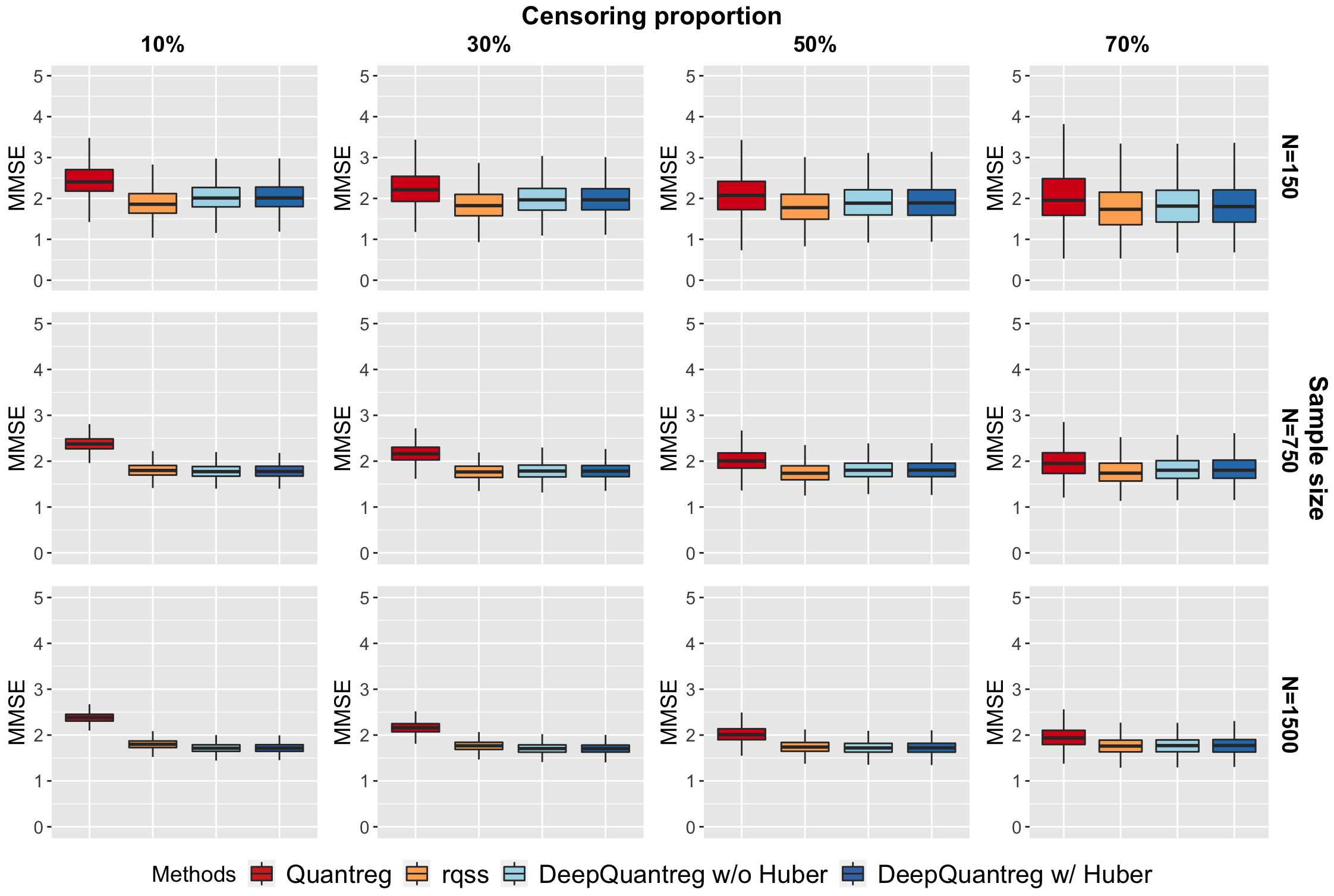}
  \caption{MMSE for group effect data}
  \label{fig:groupeffect_mmse}
\end{subfigure}

\begin{subfigure}{0.5\textwidth}
  %\centering
  % include first image
  \includegraphics[width=.95\linewidth]{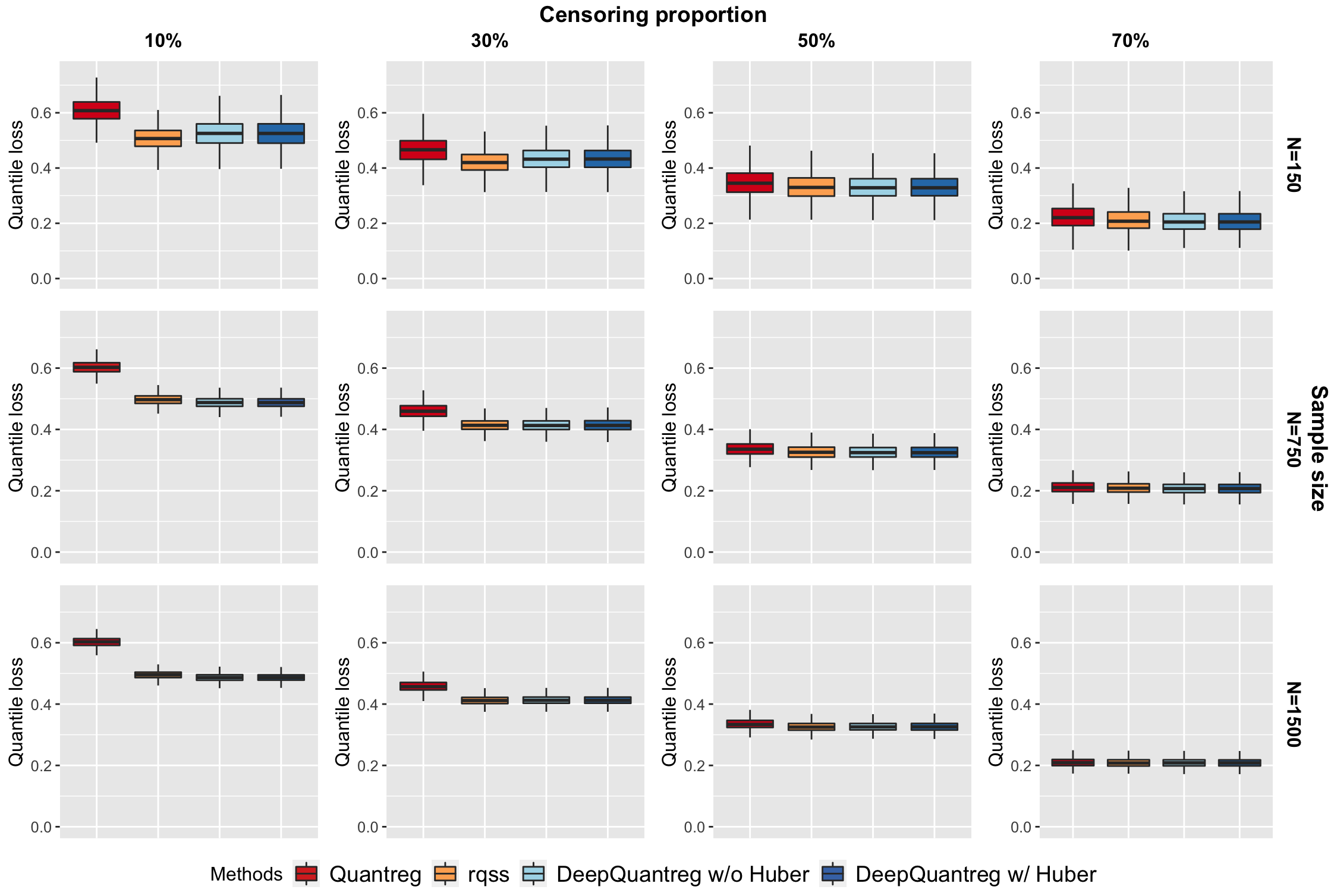}
  \caption{Quantile loss for no group effect data}
  \label{fig:noeffect_ql}
\end{subfigure}
\begin{subfigure}{0.5\textwidth}
  %\centering
  % include first image
  \includegraphics[width=.95\linewidth]{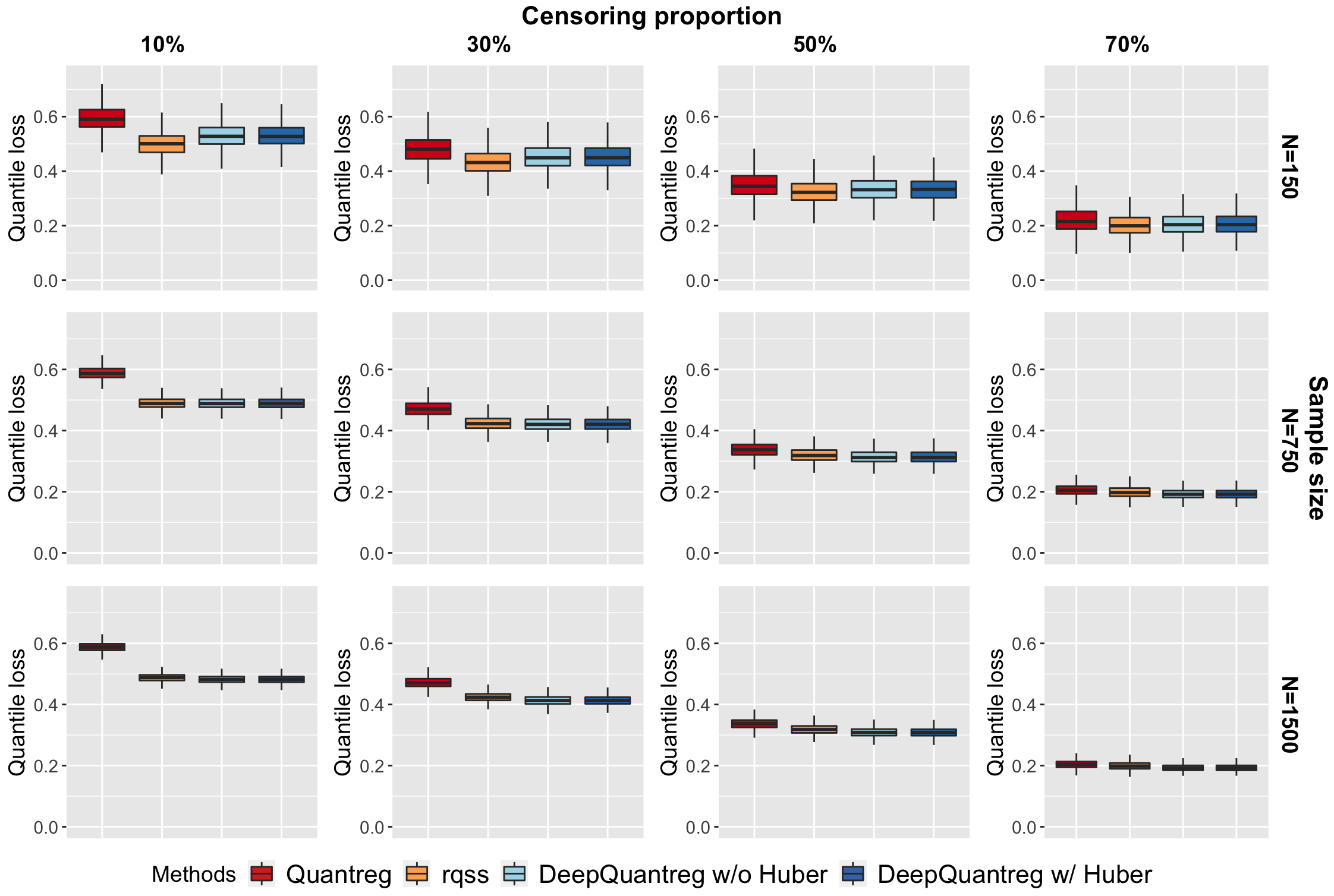}
  \caption{Quantile loss for group effect data}
  \label{fig:groupeffect_ql}
\end{subfigure}

\caption{Boxplot of $C$-Index, MMSE and quantile loss for different scenarios when $\tau$ = 0.5}
\label{fig:boxplot}
\end{figure}

First we compared performance among the models (i)-(iv) described at the beginning of this section for various sample sizes, $\tau$ values and censoring proportions. Figure \ref{fig:boxplot} shows the box plots of the $C$-index, the MMSE, and the quantile loss when $\tau$ = 0.5. The results for $\tau$ = 0.25 and 0.75 settings can be found in Supplemental Materials. One can observe that the traditional quantile regression performs the worst in all scenarios, as expected. The performance of DeepQuantreg is similar to rqss method regarding the MMSE when $\tau$ = 0.25 and 0.5, and superior to the rqss method as $\tau$ increases to 0.75. According to the quantile loss, DeepQuantreg in general outperforms the rqss method under all settings, and such superiority increases with larger $\tau$ values. On the other hand, the performance of DeepQuantreg is slightly worse than the rqss method regarding the $C$-index, especially under the small size settings. This can be also observed in Figures S1 and S2 in Supplementary Materials when $\tau=0.25$ and $\tau = 0.75$. As mentioned in Section \ref{metric}, it would be worth reiterating here that since the $C$-index is only based on the concordance rate, it might not directly measure the prediction accuracy in terms of difference between predicted and observed values as in the MMSE and the quantile loss (QL). Although we did not observe much difference in the results between DeepQuantreg with and without Huber function, we decide to still include Huber function in the loss function of DeepQuantreg in the subsequent analyses mainly due to its differentiability at the origin. 

\begin{figure}
\begin{subfigure}{0.5\textwidth}
  %\centering
  % include first image
  \includegraphics[width=.95\linewidth]{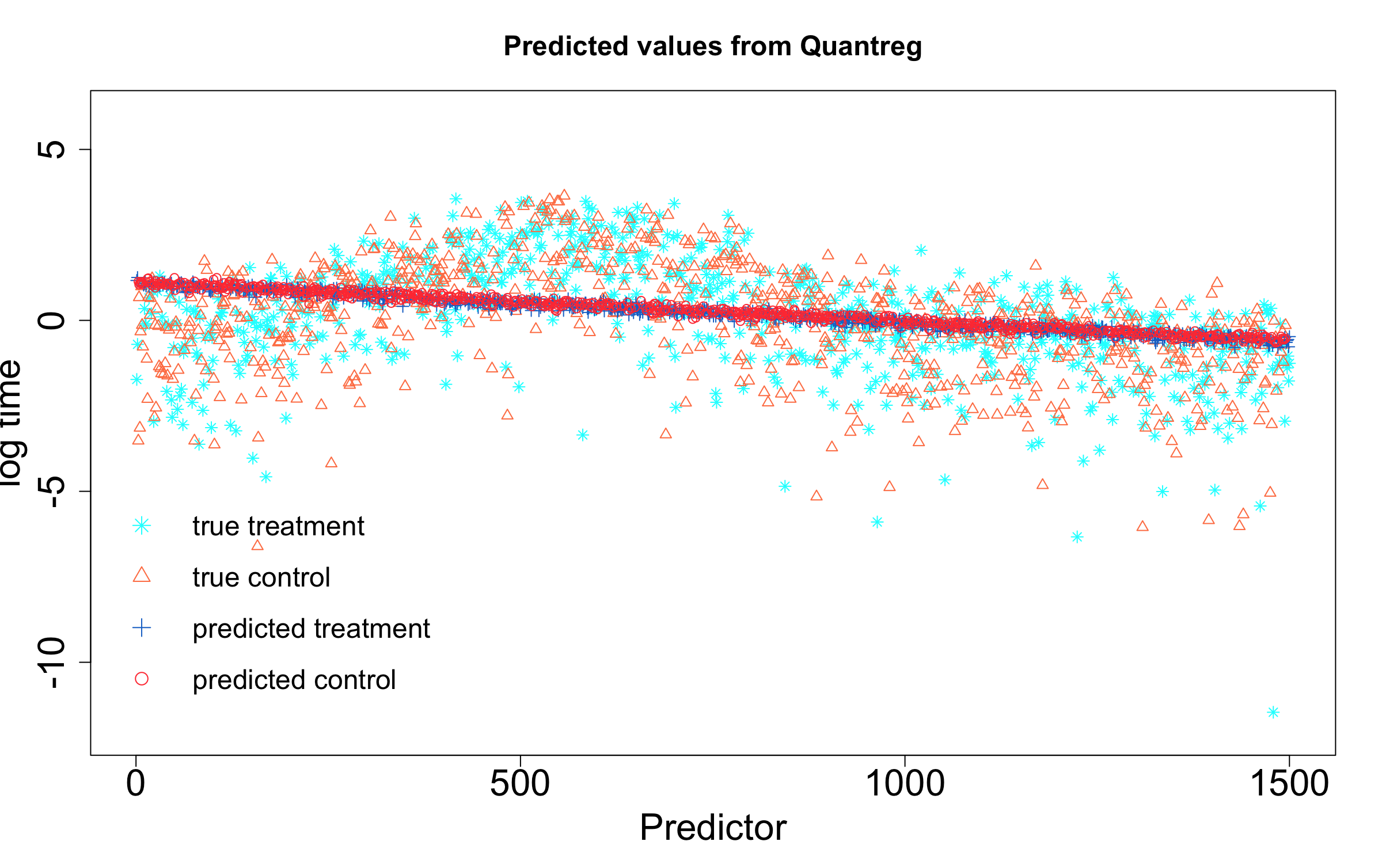}
\end{subfigure}
\begin{subfigure}{0.5\textwidth}
  %\centering
  % include second image
  \includegraphics[width=.95\linewidth]{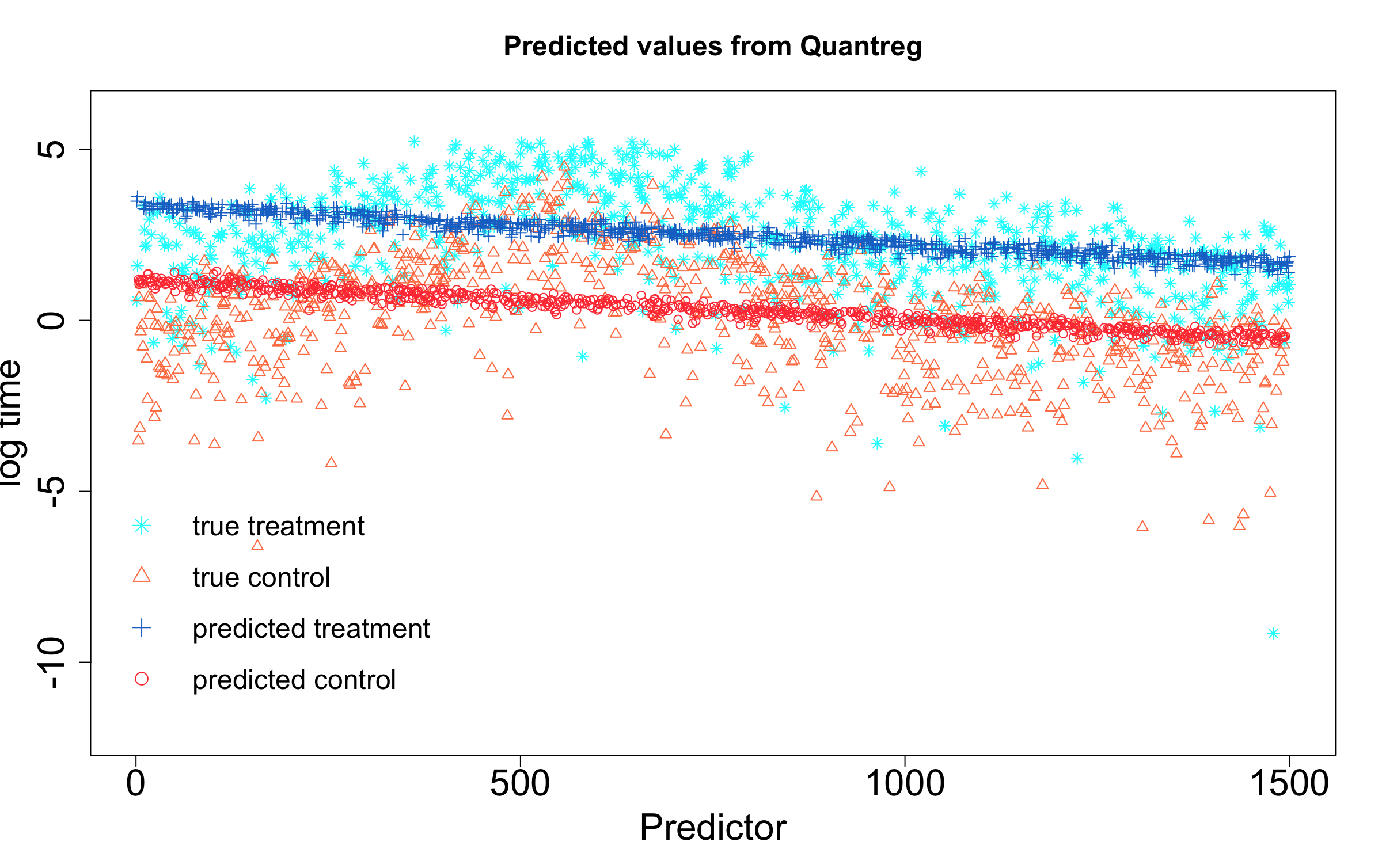}  
\end{subfigure}

\begin{subfigure}{0.5\textwidth}
  %\centering
  % include third image
  \includegraphics[width=.95\linewidth]{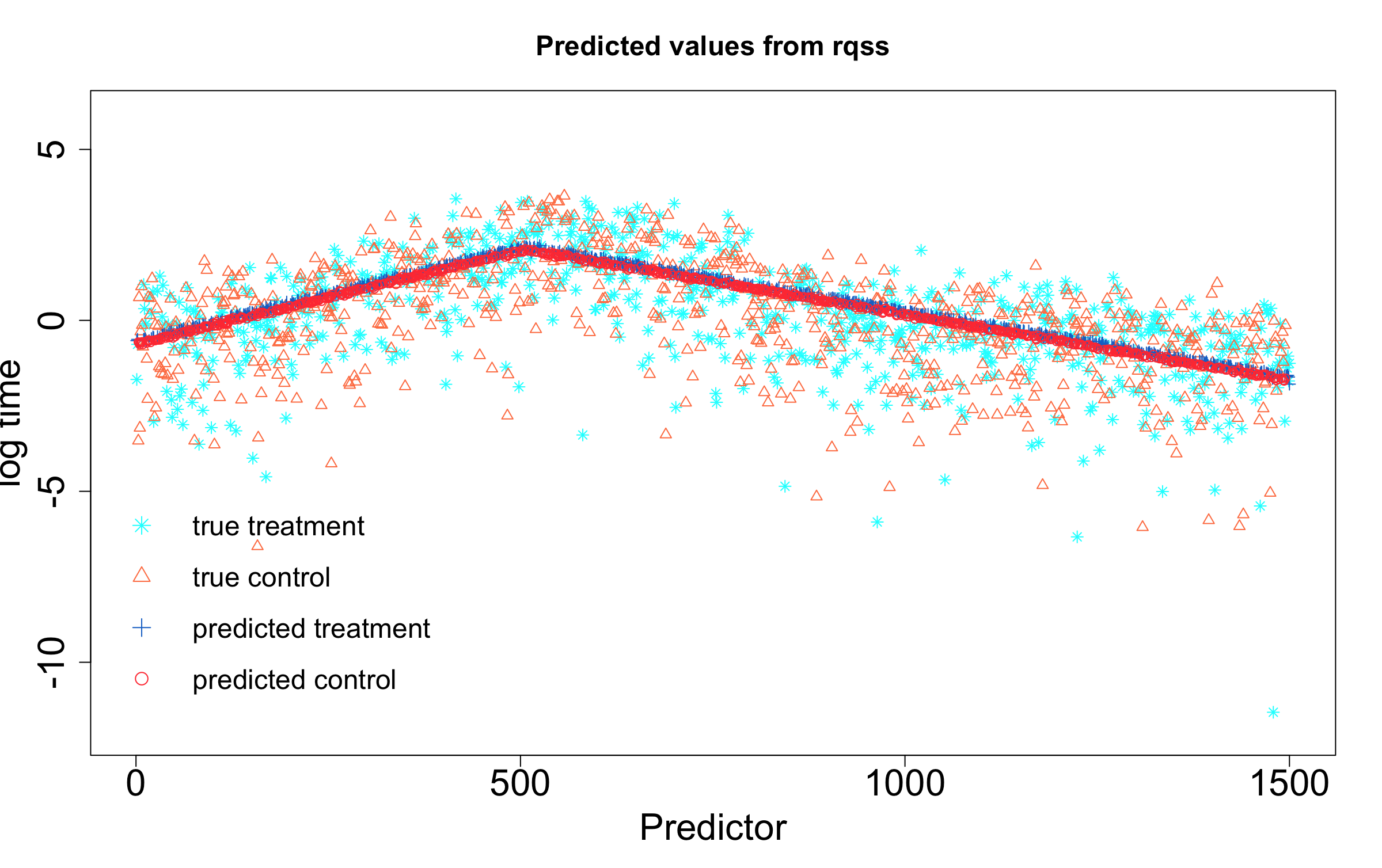}  
\end{subfigure}
\begin{subfigure}{0.5\textwidth}
  %\centering
  % include third image
  \includegraphics[width=.95\linewidth]{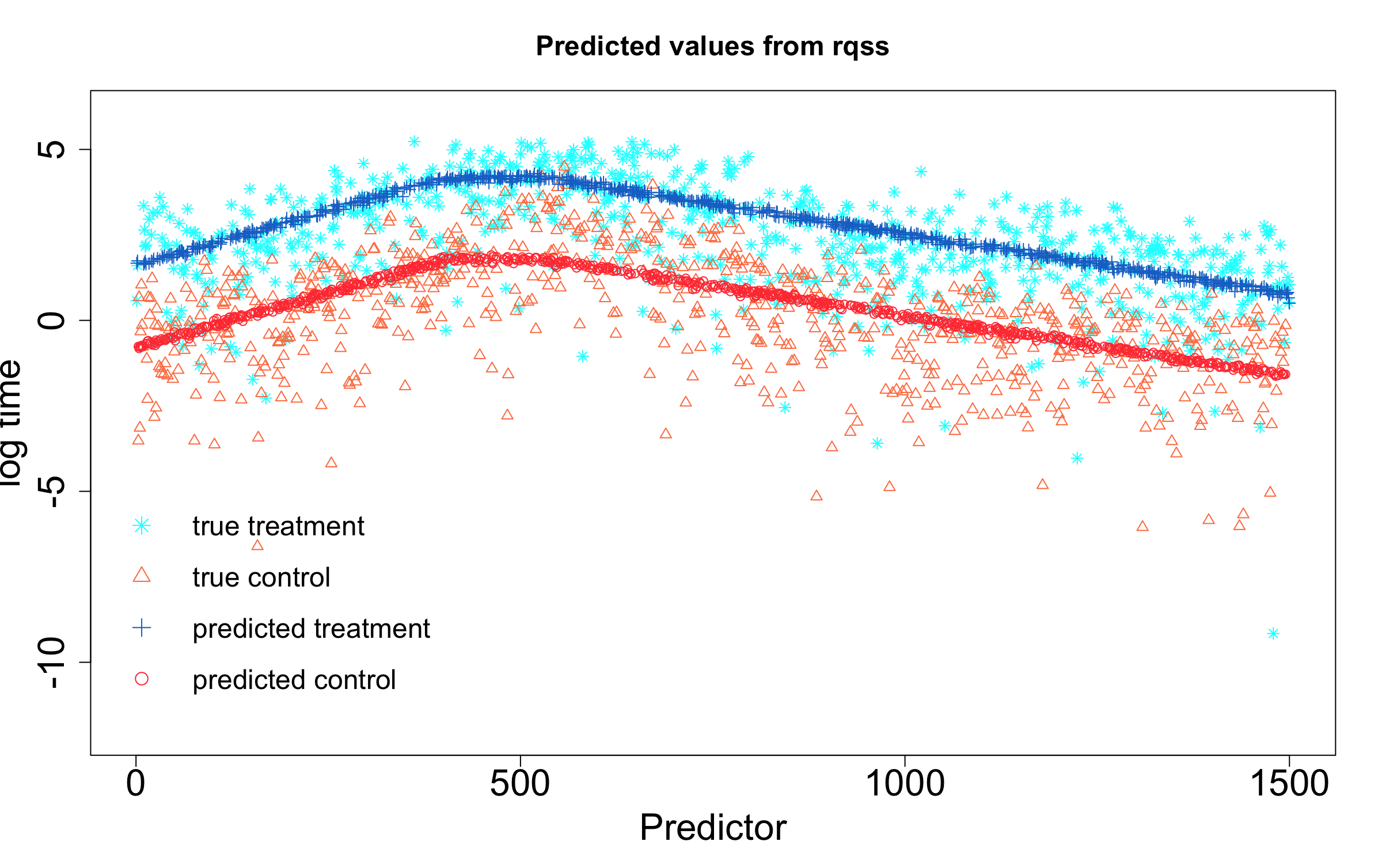}  
\end{subfigure}

\begin{subfigure}{0.5\textwidth}
  %\centering
  % include third image
  \includegraphics[width=.95\linewidth]{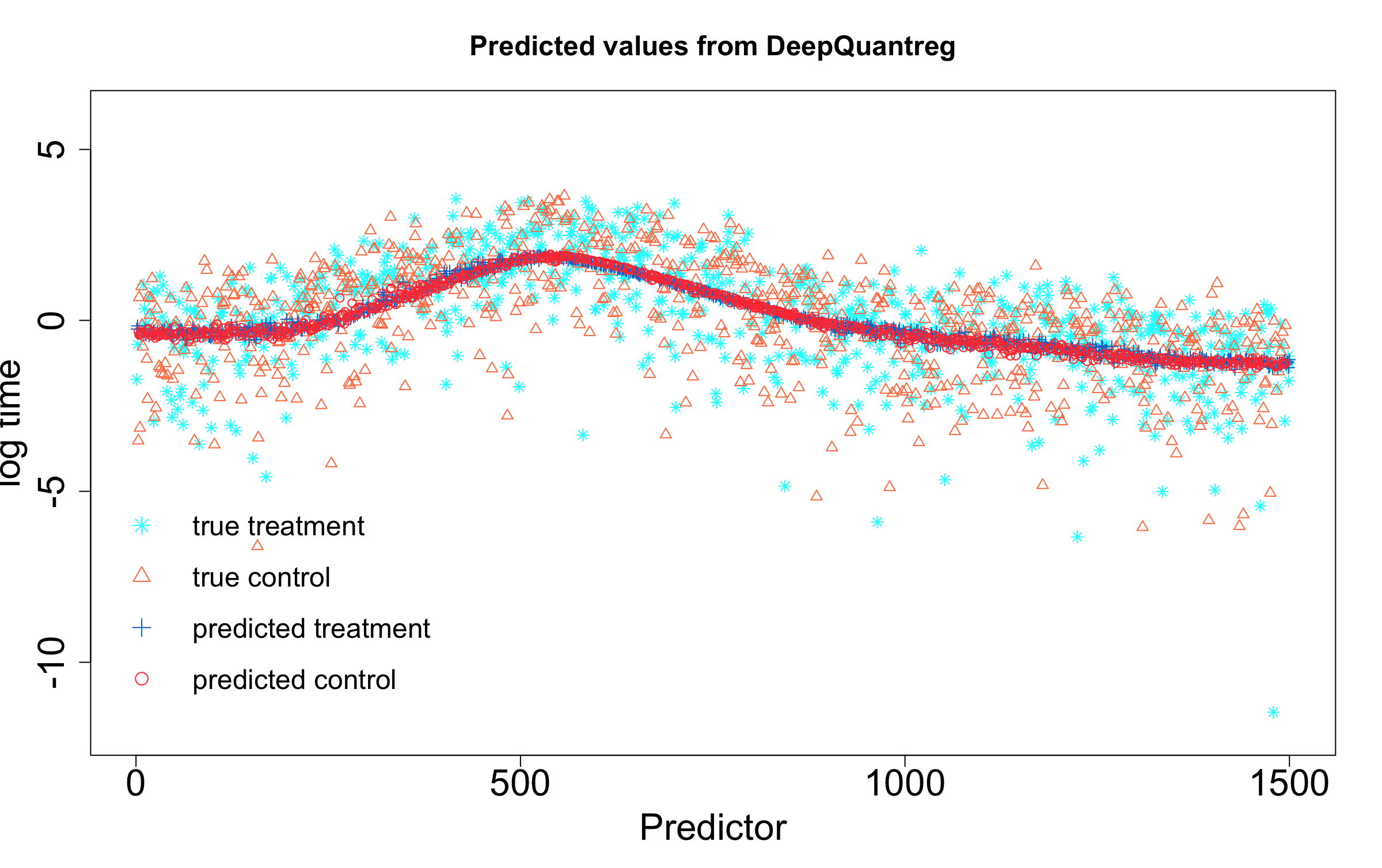}  
\end{subfigure}
\begin{subfigure}{0.5\textwidth}
  %\centering
  % include third image
  \includegraphics[width=.95\linewidth]{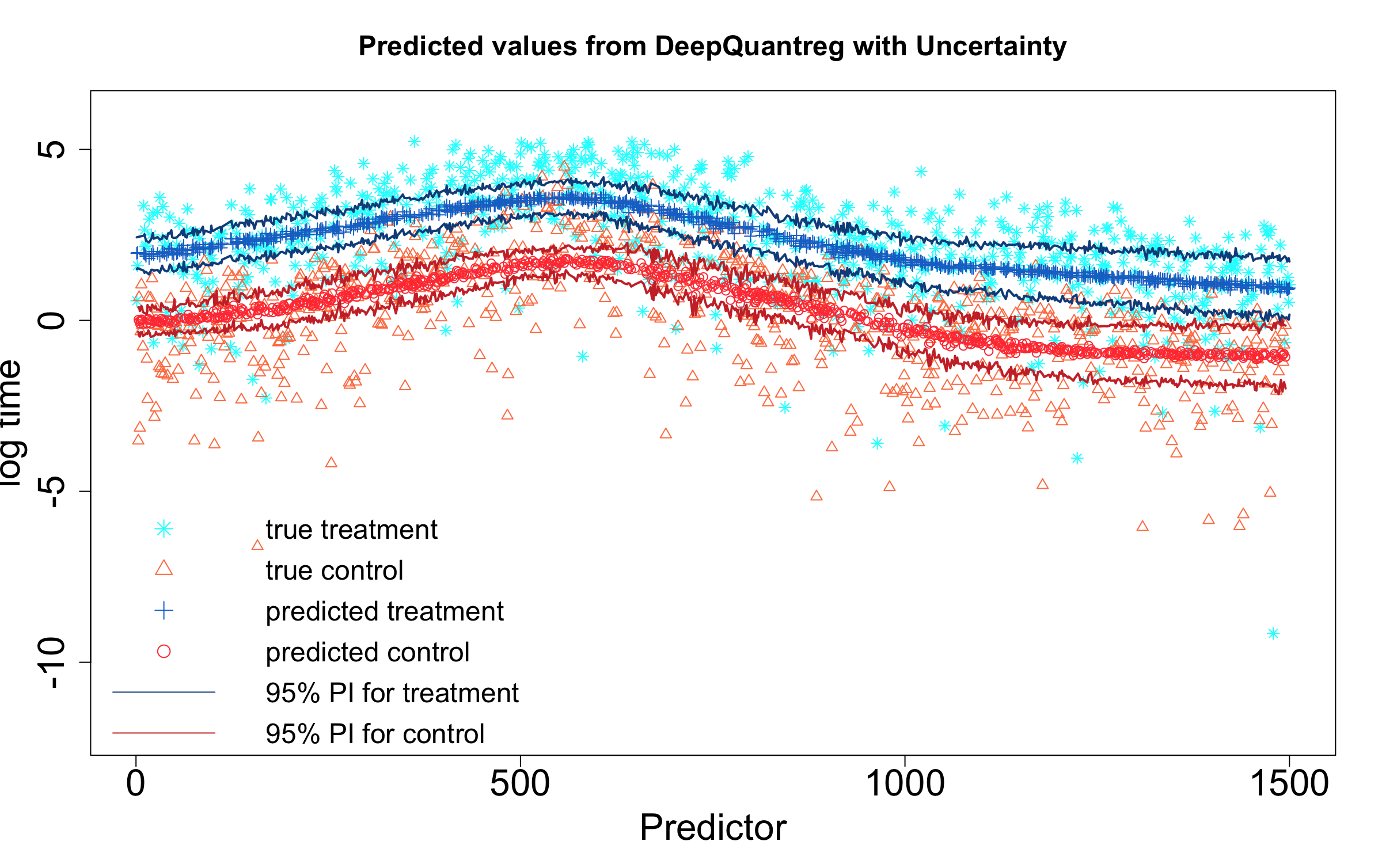}  
\end{subfigure}

\caption{Median ($\tau=0.5$) prediction plots for no group effect data (left) and group effect data (right) under 10\% censoring proportion}
\label{fig:predictionplots10}
\end{figure}

\begin{figure}
\begin{subfigure}{0.5\textwidth}
  %\centering
  % include first image
  \includegraphics[width=.95\linewidth]{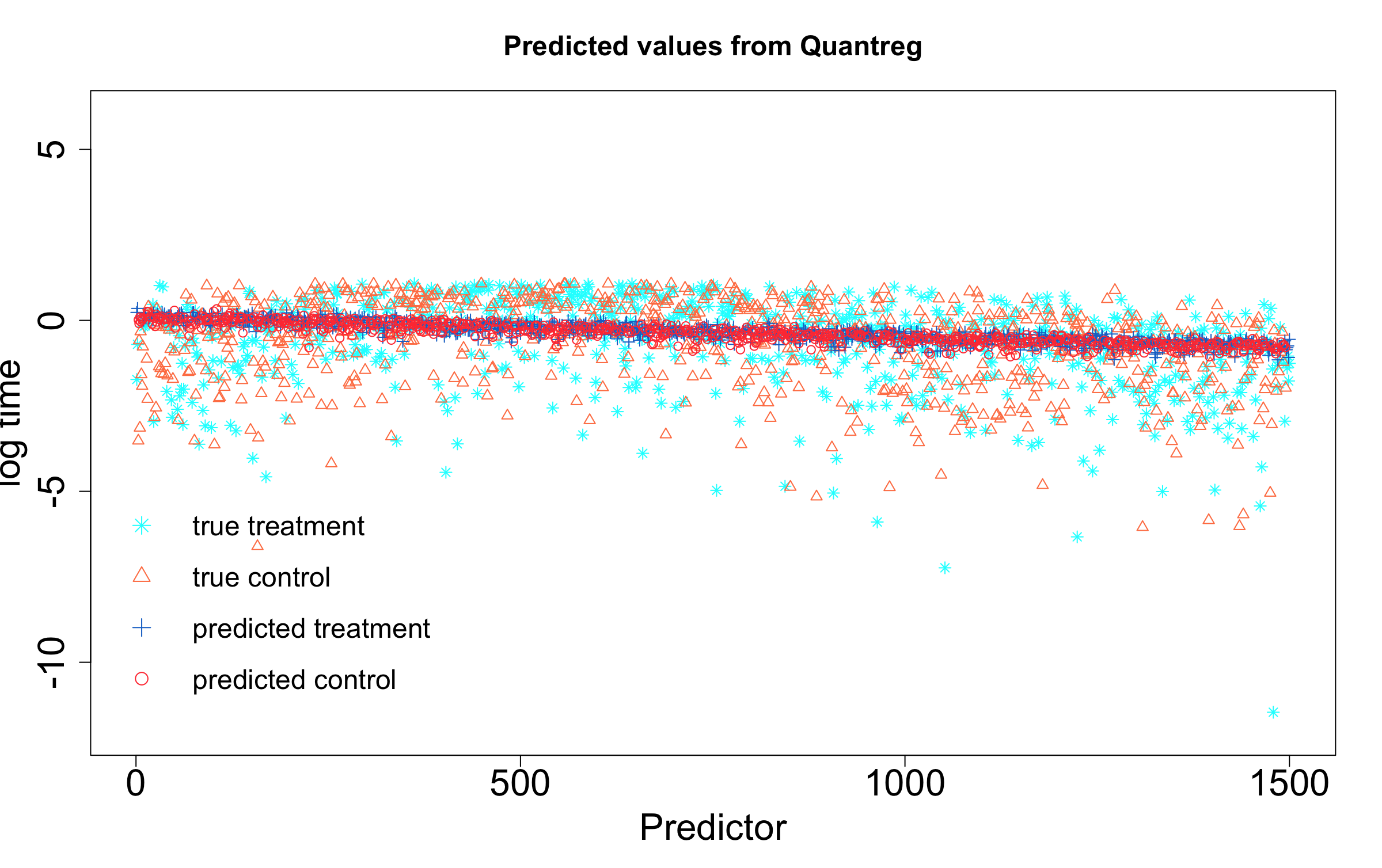}
\end{subfigure}
\begin{subfigure}{0.5\textwidth}
  %\centering
  % include second image
  \includegraphics[width=.95\linewidth]{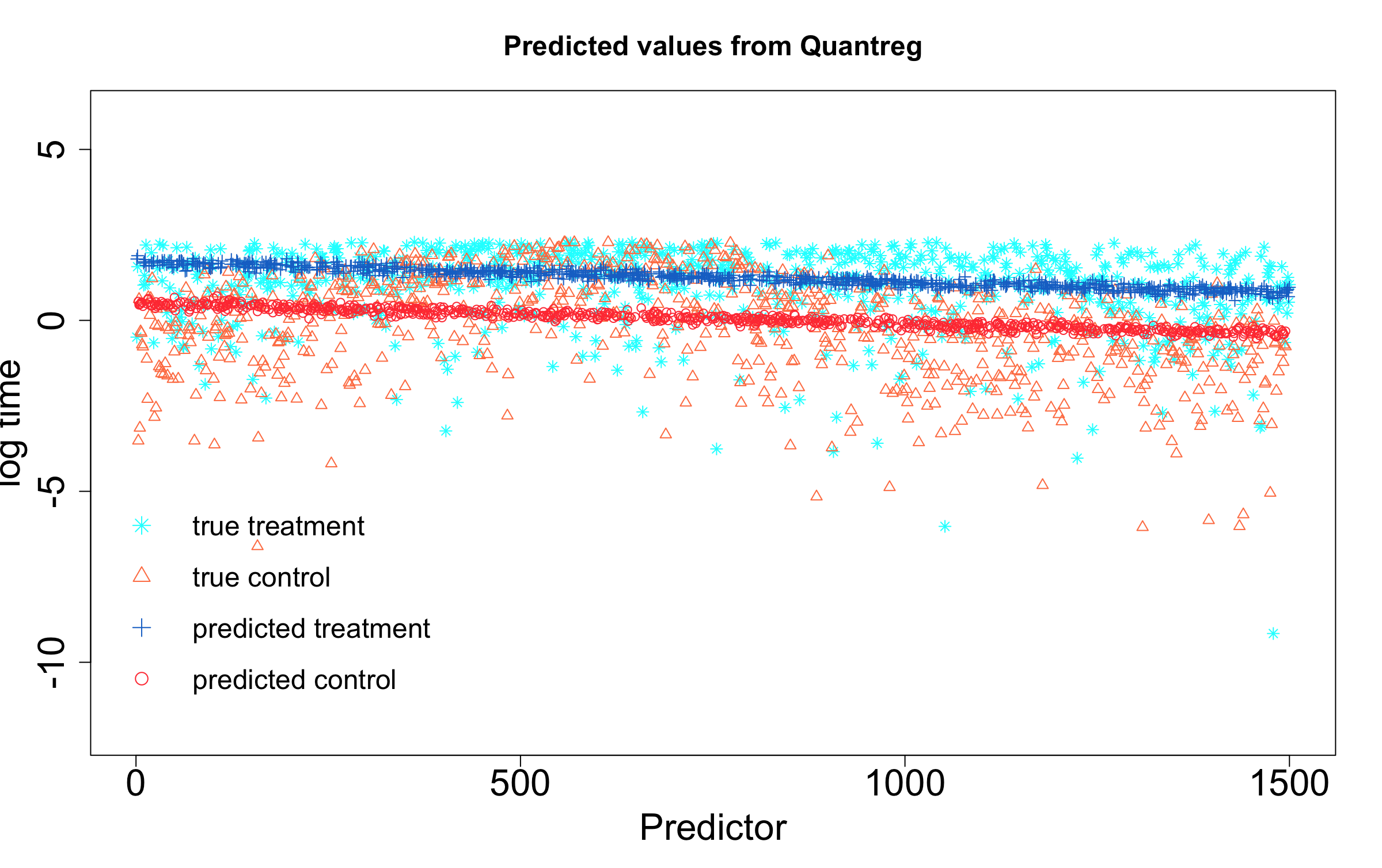}  
\end{subfigure}

\begin{subfigure}{0.5\textwidth}
  %\centering
  % include third image
  \includegraphics[width=.95\linewidth]{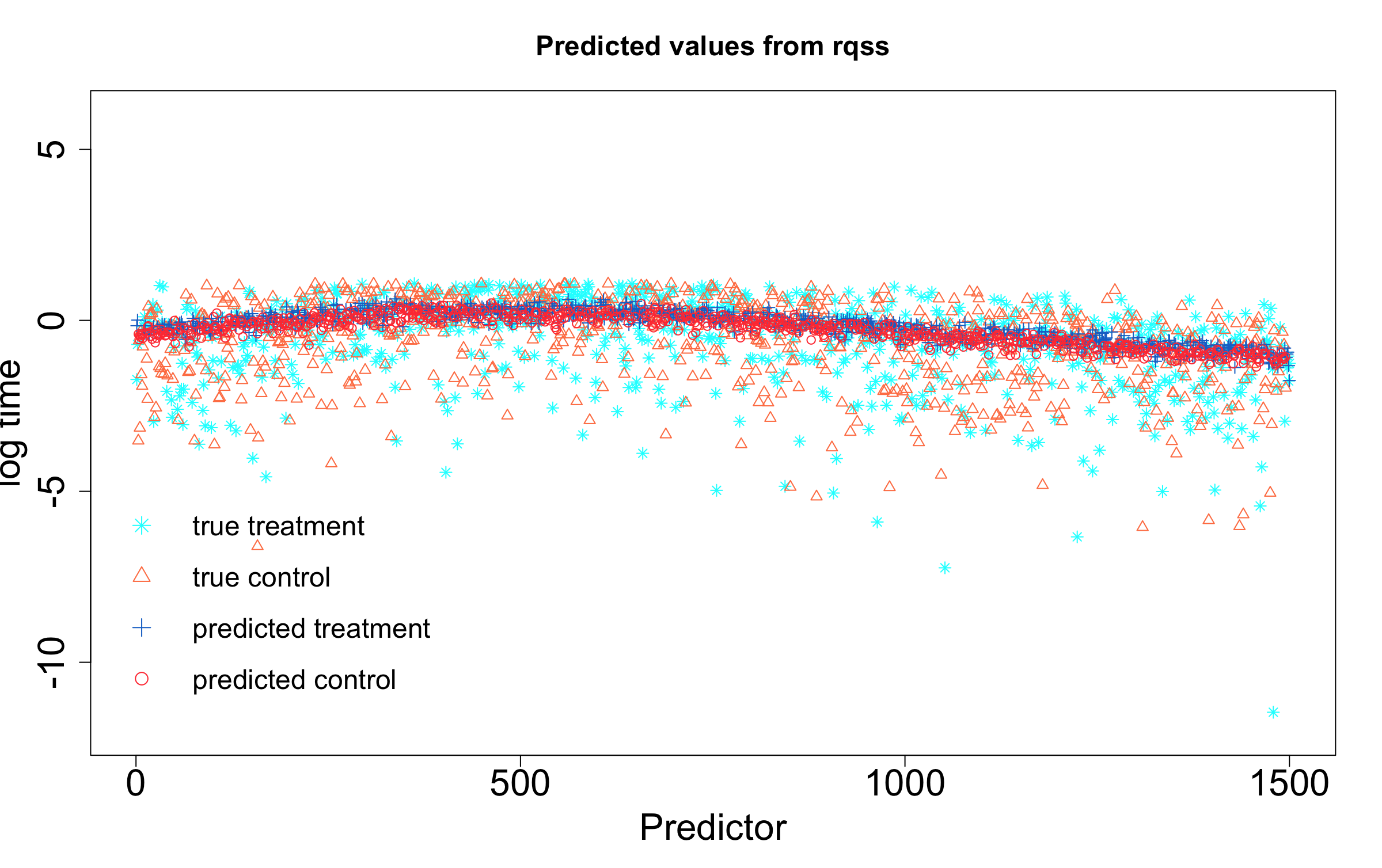}  
\end{subfigure}
\begin{subfigure}{0.5\textwidth}
  %\centering
  % include third image
  \includegraphics[width=.95\linewidth]{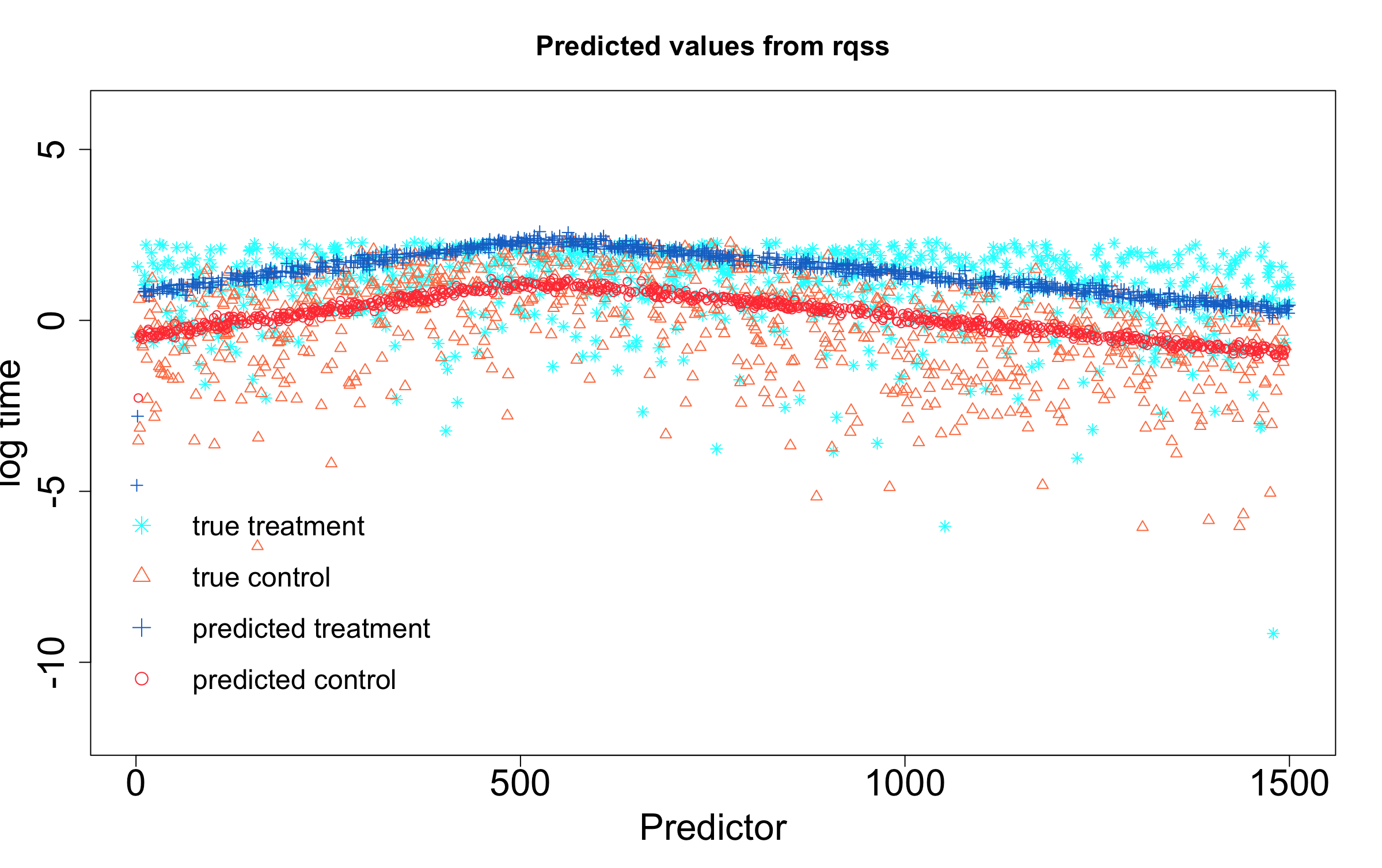}  
\end{subfigure}

\begin{subfigure}{0.5\textwidth}
  %\centering
  % include third image
  \includegraphics[width=.95\linewidth]{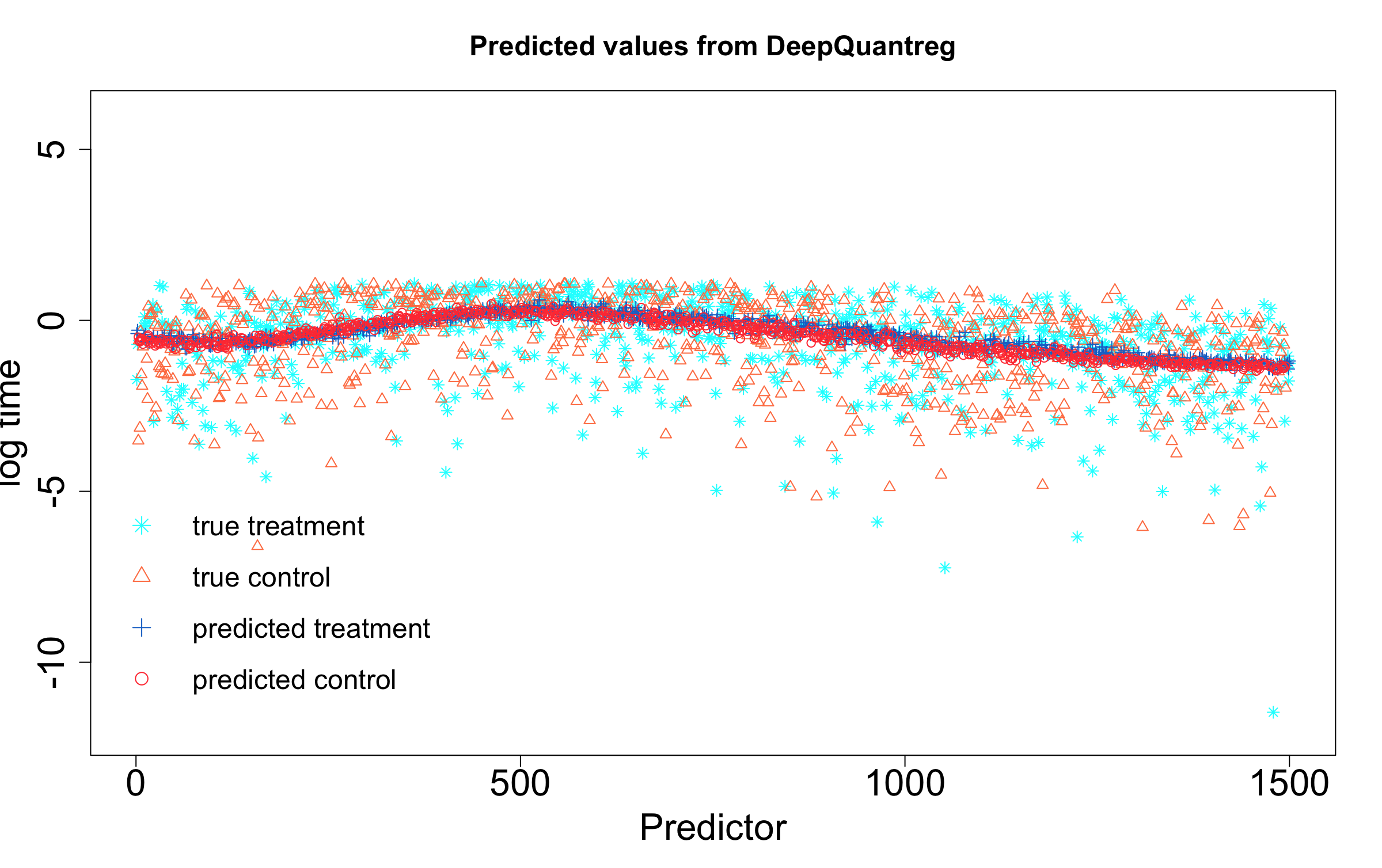}  
\end{subfigure}
\begin{subfigure}{0.5\textwidth}
  %\centering
  % include third image
  \includegraphics[width=.95\linewidth]{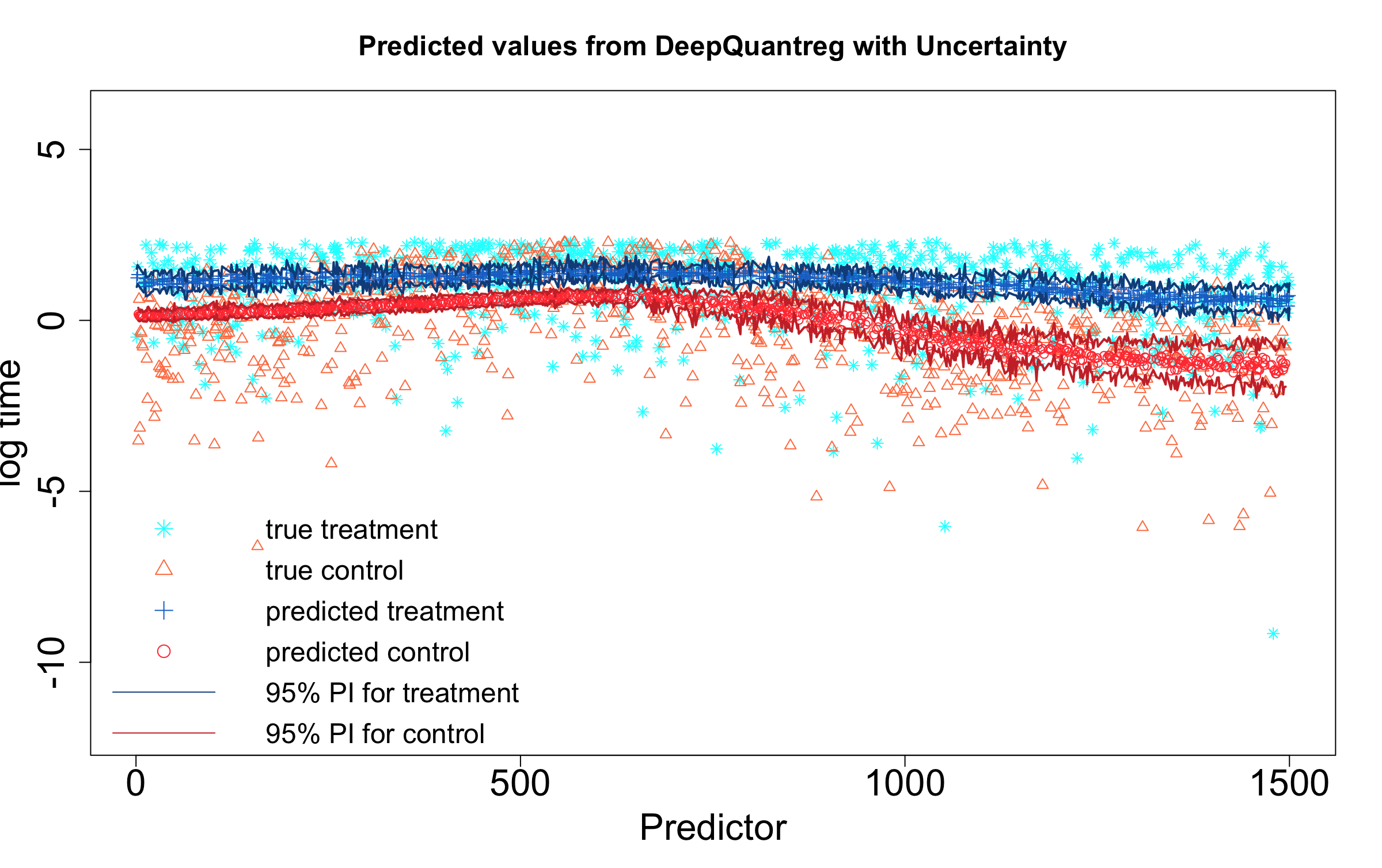}  
\end{subfigure}

\caption{Median ($\tau=0.5$) prediction plots for the no group effect data (left) and group effect data (right) under 50\% censoring proportion}
\label{fig:predictionplots50}
\end{figure}

In Figures 4 and 5, we also present the median ($\tau = 0.5$) prediction results on the test sets with censoring proportions of 10\% and 50\% for the no group effect data and group effect data, together with 95\% prediction intervals from DeepQuantreg only for the group effect data using the dropout method described in Section \ref{pi}. Figure \ref{fig:predictionplots10} shows the fitting results when the censoring proportion is 10\%, where $\circ$ indicates the fitted values for the control group and $+$ for the intervention group, implying that DeepQuantreg and the rqss method captures the nonlinear patterns in the data reasonably well in both scenarios. Although the same check function was used in the loss function, the traditional quantile regression model fails to capture the nonlinear patterns due to the restricted linear assumption. Figure \ref{fig:predictionplots50} shows a similar plot for the case of 50\% censoring. One can observe that DeepQuantreg also captures the nonlinear patterns appropriately under heavy censoring, demonstrating the ability of capturing different patterns for different groups in this example, i.e. a unimodal shape for control group and a flatter shape for intervention group due to the censoring effect of the larger true event times in the middle of the curve. Even though the $C$-index from the rqss method is higher than one from DeepQuantreg in Figure \ref{fig:boxplot}, Figure \ref{fig:predictionplots50} indicates that the rqss procedure fails to capture different shapes in different groups since the \texttt{qss()} function cannot be applied to categorical variables, which seems one of the limitations of the rqss method, along with some challenges in choosing the smoothing parameter $\lambda$ \citep{koenker2011additive}. This also explains the increased superiority in the MMSE and QL of DeepQuantreg with higher quantiles (Figures S1 and S2 in Supplementary Materials) because the shapes of curves corresponding to the two groups tend to be more different at higher quantiles.

\subsection{Effects of Number of Nodes}
We evaluated the effect of the number of hidden layers and the number of nodes per layer on the performance of the proposed deep censored median regression model, i.e. when $\tau=0.5$, for the group effect data set. Supplemental Figure S3a shows that the model using one hidden layer with 4 nodes can only predict a linear fit. However, increasing the number of layers seems to help capturing the non-linearity. Supplemental Figure S3b shows that with 300 nodes per layer, even one hidden layer can approximate the non-linearity reasonably well, and increasing the number of layers will further sharpen the nonlinear fit. In many studies using deep neural network, the number of nodes in hidden layers tends to decrease toward the output, being less than or equal to the number of input variables. Therefore, our finding here interestingly indicates that given the same number of layers the neural network predicts the nonlinear patterns better when the number of nodes in the hidden layer is larger than the number of input variables.

\section{Real Data Application}
\label{s:data}
In this section, we compared DeepQuantreg with traditional and nonparametric quantile regression with the Netherlands Cancer Institute 70 gene signature data set \citep{van2002gene}, which is publicly available from the R package \verb!penalized! and will be referred to as NKI70 data set, and the Molecular Taxonomy of Breast Cancer International Consortium (METABRIC) data set \citep{curtis2012genomic}. 

The NKI70 data set contains 144 lymph node positive breast cancer patients' information on metastasis-free survival, 5 clinical risk factors, and gene expression measurements of 70 genes found to be prognostic for metastasis-free survival in an earlier study. The censoring proportion is around 67\%. The covariates we included in the model are age at diagnosis, ER status, and 6 genes that show nonlinear relationship with time to metastasis or last follow-up (Supplemental Figure S4). 

METABRIC data set consists of 1981 breast cancer patients' gene expression data and 21 clinical features. We included 4 gene indicators (MKI67, EGFR, PGR, and ERBB2) with 5 clinical features (ER, HER2 and PR status, tumor size, and age at diagnosis). We restricted to the 1884 patients with complete data on the included covariates, where the censoring proportion is 42.14\%.

Both of the data sets were randomly splitted into $\frac{2}{3}$ training and $\frac{1}{3}$ test sets and repeated for 50 times to calculate the average performance based on the $C$ -index, MMSE and quantile loss. Note that we did the hyperparameter tuning in each of the 50 training sets separately.

Since the nonparametric quantile regresion method cannot extrapolate during prediction, Table \ref{tab:NKI70} and Table \ref{tab:METABRIC} show the prediction results for the subsets of test sets in which the covariate values are within the range of the values in the training set for all three methods. In case of the NKI70 data set, as shown in Table \ref{tab:NKI70}, the prediction performance of the traditional and nonparametric quantile regression generally decreases as the quantile increases while our deep learning algorithm stays stable over all three quantiles. One plausible explanation is that the data are not as non-linear at the $25^{th}$ quantile compared to higher quantiles, so that DeepQuantreg outperforms the other two methods only at the higher quantiles. In the analysis of the NKI70 data set, one can also observe that the proposed deep learning algorithm performs reasonably well even when the sample size is small/moderate and the number of covariates is small. For METABRIC data (Table \ref{tab:METABRIC}) with a larger sample size, DeepQuantreg outperforms Quantreg and the rqss method in all three metrics, as expected.

\begin{table}
\centering
\caption{$C$-index, MMSE and quantile loss (mean(SD)) of the NKI70 data under different quantiles $\tau$} 
\label{tab:NKI70}
\begin{adjustbox}{width=0.8\textwidth}
\begin{tabular}{ccccc}
\hline
Metric        & Quantile  & Quantreg          & rqss            & DeepQuantreg   \\
\hline
              & 0.25 & 0.723 (0.073)     & 0.734 (0.056)   &  0.734 (0.060)   \\
C-index       & 0.5  & 0.670 (0.098)      & 0.687 (0.068)   & 0.735 (0.058)  \\
              & 0.75 & 0.603 (0.110)      & 0.644 (0.077)   & 0.734 (0.059)  \\
\hline
              & 0.25 & 0.891 (0.650)     & 0.492 (0.168)   & 0.948 (0.307) \\
MMSE          & 0.5  & 1.253 (0.833)  & 0.888 (0.283)  & 0.949 (0.311) \\
              & 0.75 & 1.741 (0.774) & 1.589 (0.415) & 0.949 (0.306) \\
\hline
              & 0.25 & 0.306 (0.077)     & 0.882 (0.184)   & 0.222 (0.059)  \\
Quantile loss & 0.5  & 0.386 (0.090)     & 1.750 (0.727)   & 0.196 (0.055)   \\
              & 0.75 & 0.327 (0.052)     & 0.999 (0.307)   & 0.168 (0.069)\\
\hline
\end{tabular}
\end{adjustbox}
\end{table}

\begin{table}
\centering
\caption{$C$-index, MMSE and quantile loss (mean(SD)) of the METABRIC data under different quantiles $\tau$} 
\label{tab:METABRIC}
\begin{adjustbox}{width=0.8\textwidth}
\begin{tabular}{ccccc}
\hline
Metric        & Quantile  & Quantreg          & rqss            & DeepQuantreg   \\
\hline
              & 0.25 & 0.613 (0.021)        & 0.632 (0.020)       & 0.657 (0.019)      \\
C-index       & 0.5  & 0.592 (0.020)         & 0.648 (0.018)      & 0.657 (0.019)      \\
              & 0.75 & 0.571 (0.023)        & 0.627 (0.021)      & 0.657 (0.018)      \\
\hline
              & 0.25 & 0.794 (0.123)    & 0.782 (0.091) & 0.779 (0.123) \\
MMSE          & 0.5  & 0.988 (0.151)  & 0.788 (0.121) & 0.781 (0.120) \\
              & 0.75 & 1.848 (0.224) & 1.339 (0.189)  & 0.780 (0.118) \\
\hline
               & 0.25 & 3.242 (0.898)       & 3.352 (2.215)     & 0.314 (0.033)     \\
Quantile loss & 0.5  & 2.451 (0.566)       & 2.310 (0.523)     & 0.329 (0.041)     \\
              & 0.75 & 1.365 (0.285)       & 1.256 (0.257)      & 0.342 (0.068)   \\
\hline
\end{tabular}
\end{adjustbox}
\end{table}

\section{Discussion}
\label{s:discuss}

In this paper, we developed a deep learning algorithm for the quantile regression under right censoring. We adopted the Huber check function in the loss function with inverse probability weights to adjust for censoring. In addition, we illustrated how the number of hidden layers and the number of hidden nodes per layer affect the predication ability of the proposed algorithm. Our finding is that to capture the nonlinear patterns properly, one may need more hidden layers and/or more nodes per layer than the number of input variables. 

Typical approaches for regularization to prevent overfitting in machine learning include using a penalty function, and/or adding dropout layers as in neural network, both of which are readily available in our source code, newly featured with prediction intervals. To reduce the training time, our package can also utilize graphics processing units (GPU), which is freely available on cloud servers such as Google Colab.

\section*{Acknowledgments}
This research was supported in part by the University of Pittsburgh Center for Research Computing through the resources provided.

\newpage

\bibliographystyle{biom}
%\bibliography{mybib}

\newpage

\section*{Supplementary Material}

\setcounter{table}{0}
\renewcommand{\thetable}{S\arabic{table}}

\setcounter{figure}{0}
\renewcommand{\thefigure}{S\arabic{figure}}

\begin{figure}[h]
\begin{subfigure}{0.5\textwidth}
  \centering
  % include first image
  \includegraphics[width=.9\linewidth]{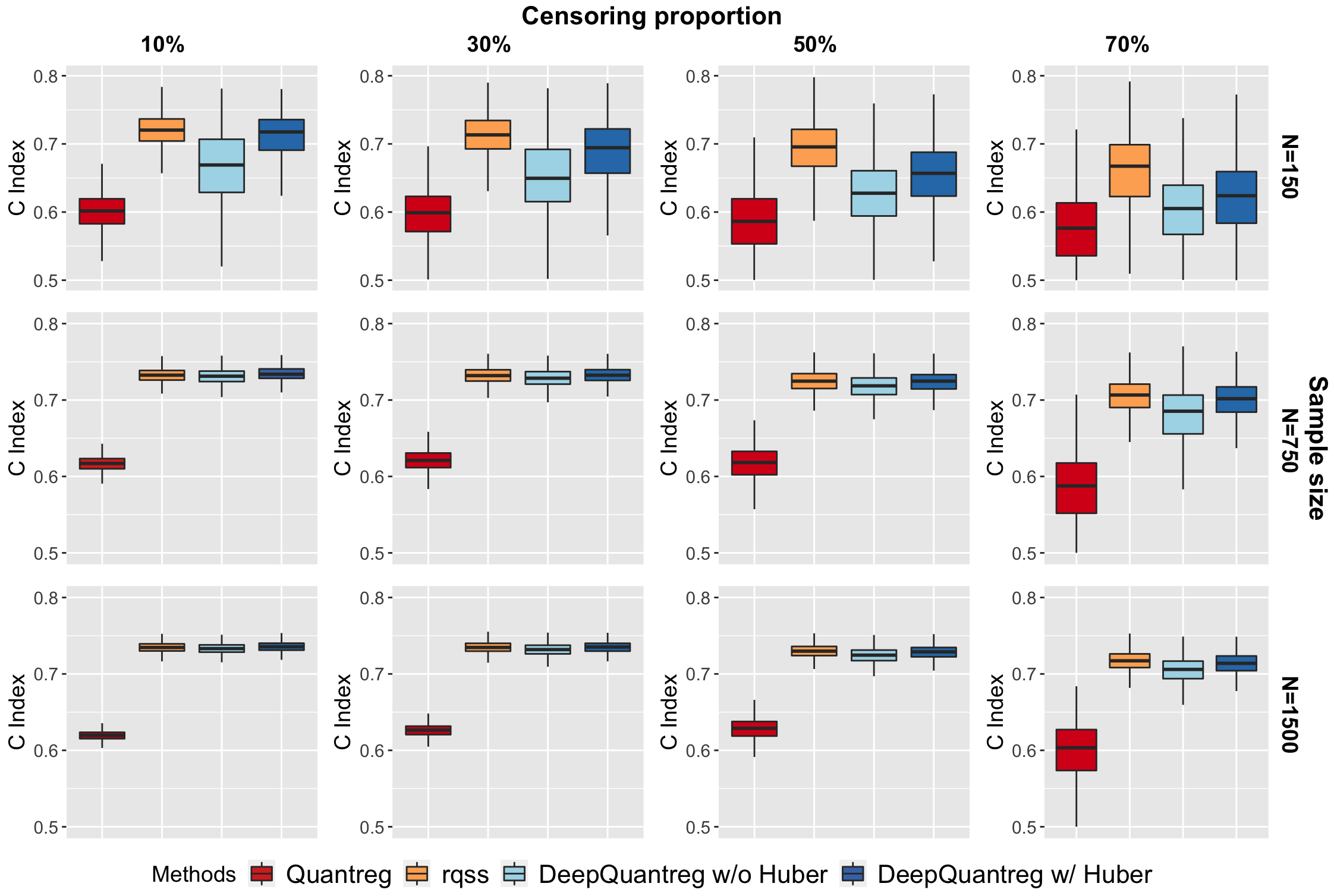}
  \caption{$C$-Index for no group effect data}
  \label{fig:noeffect_cindex}
\end{subfigure}
\begin{subfigure}{0.5\textwidth}
  \centering
  % include first image
  \includegraphics[width=.9\linewidth]{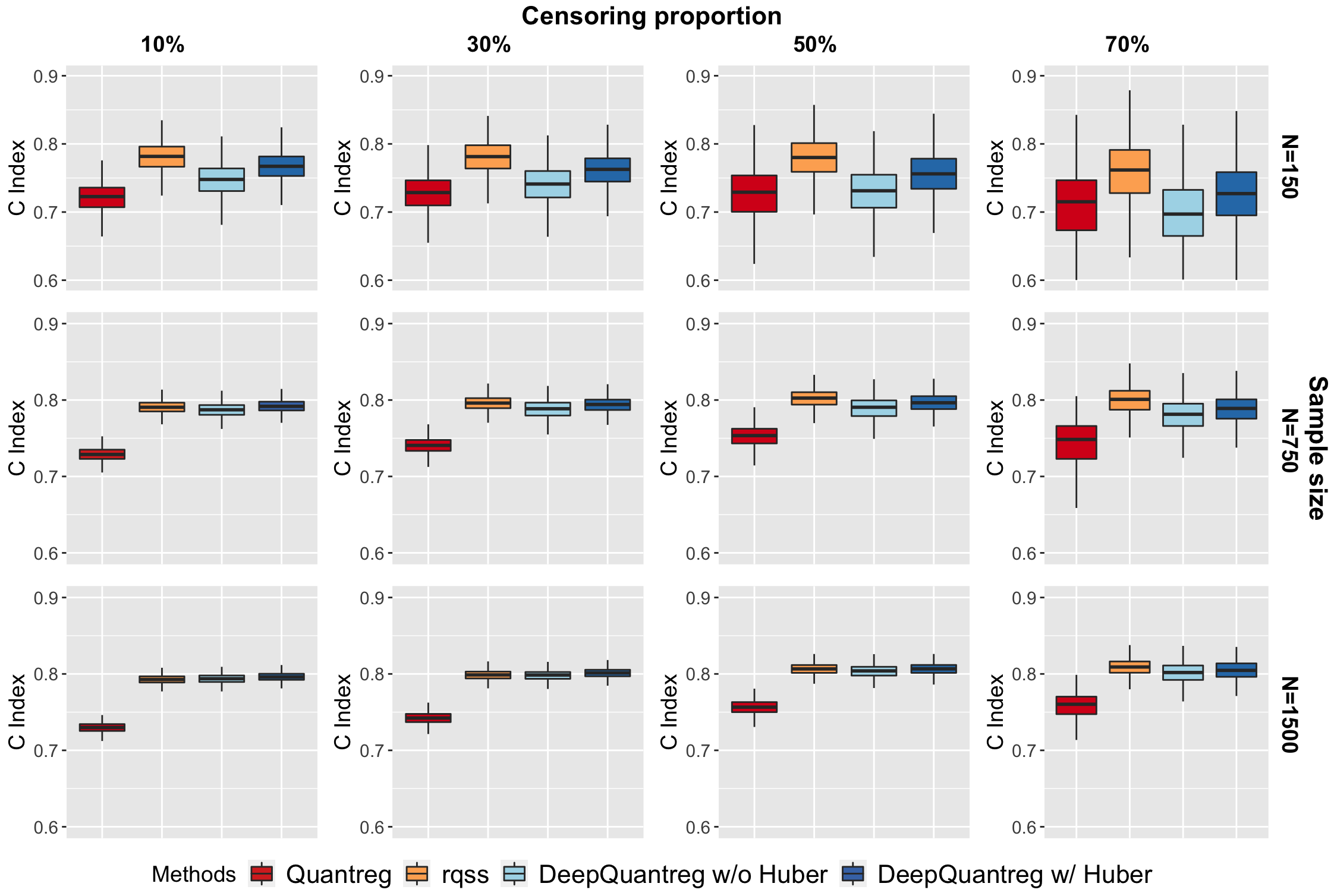}
  \caption{$C$-Index for group effect data}
  \label{fig:groupeffect_cindex}
\end{subfigure}

\begin{subfigure}{0.5\textwidth}
  \centering
  % include second image
  \includegraphics[width=.9\linewidth]{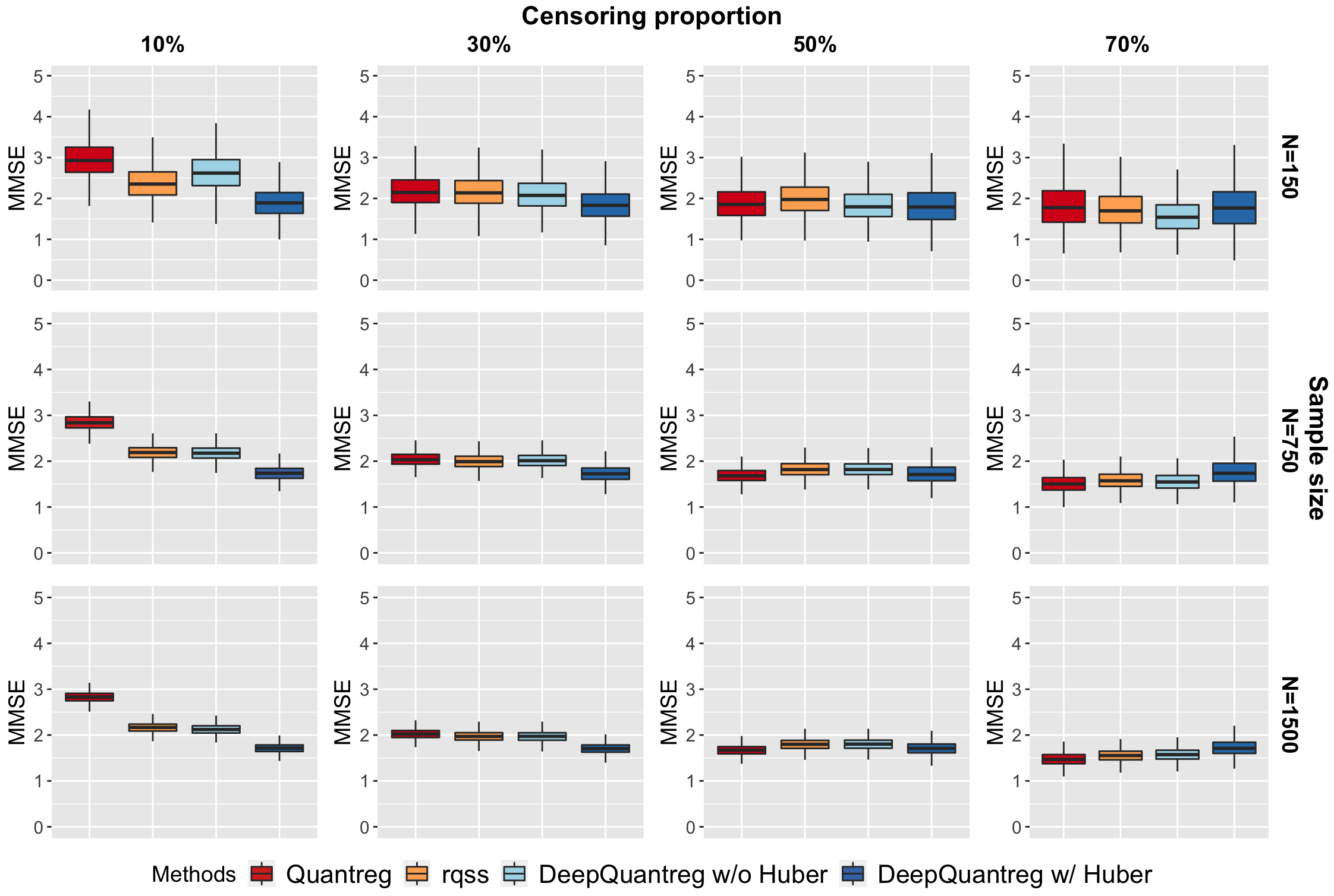}  
  \caption{MMSE for no group effect data}
  \label{fig:noeffect_mmse}
\end{subfigure}
\begin{subfigure}{0.5\textwidth}
  \centering
  % include first image
  \includegraphics[width=.9\linewidth]{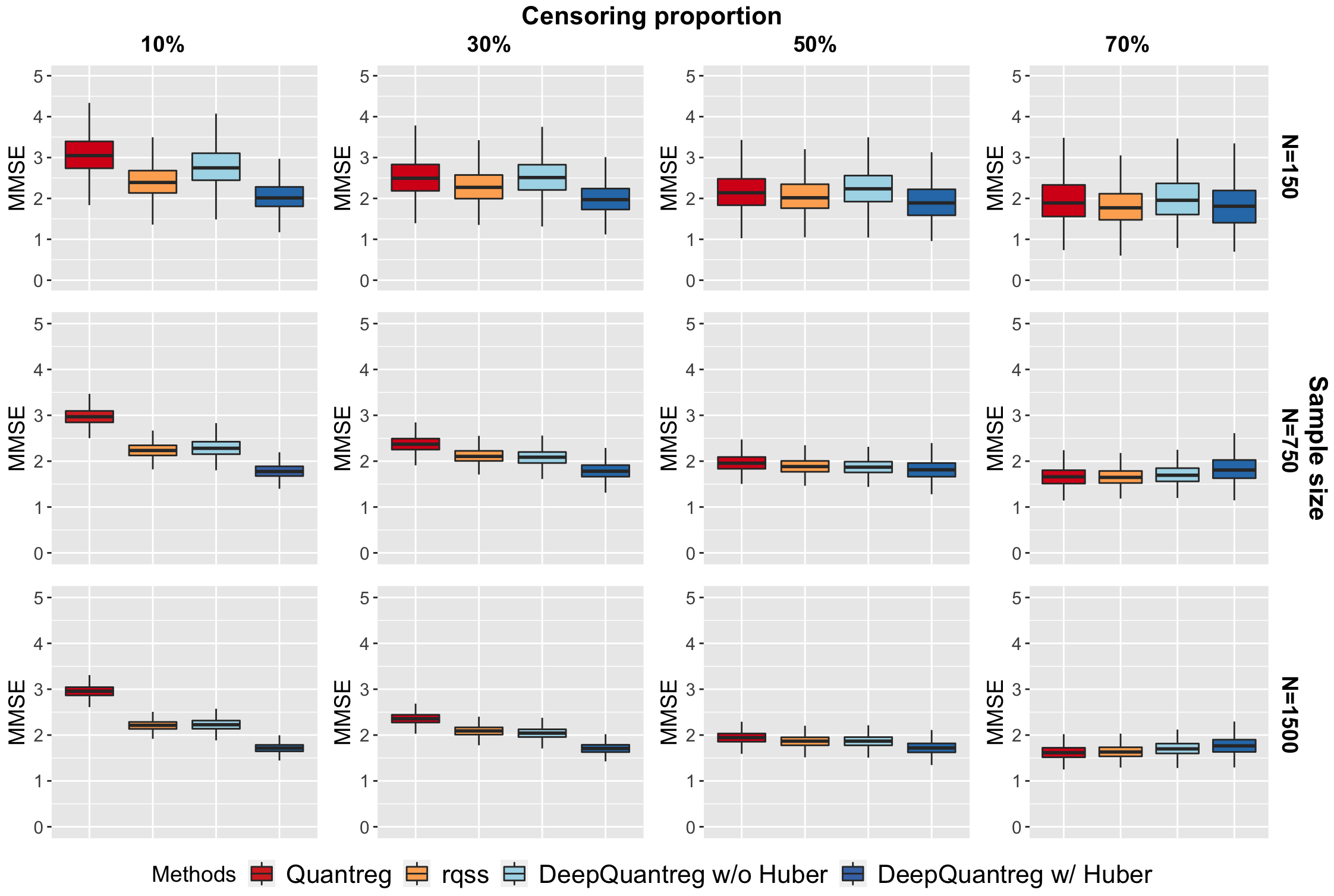}
  \caption{MMSE for group effect data}
  \label{fig:groupeffect_mmse}
\end{subfigure}

\begin{subfigure}{0.5\textwidth}
  \centering
  % include first image
  \includegraphics[width=.9\linewidth]{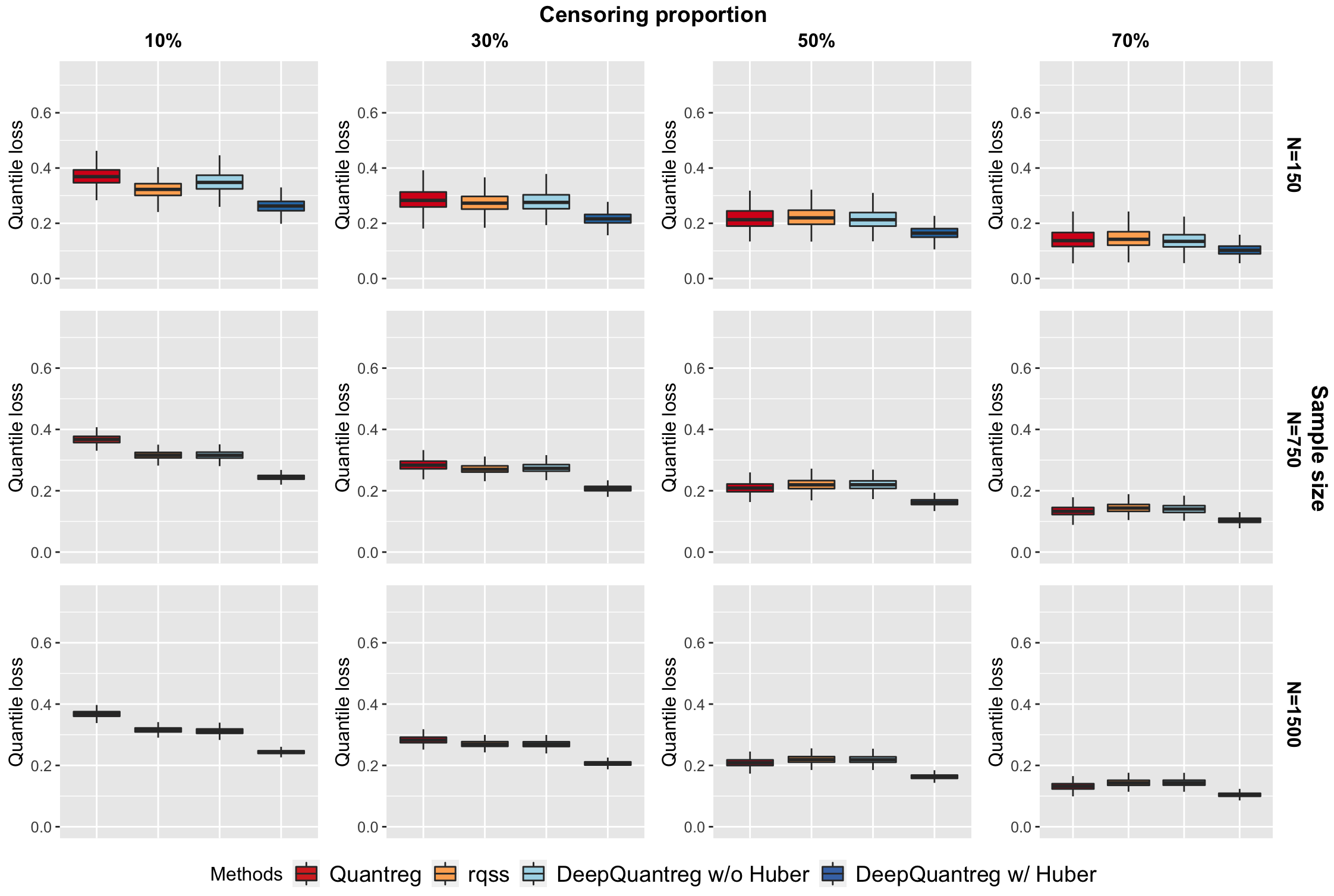}
  \caption{QL for no group effect data}
  \label{fig:noeffect_ql}
\end{subfigure}
\begin{subfigure}{0.5\textwidth}
  \centering
  % include first image
  \includegraphics[width=.9\linewidth]{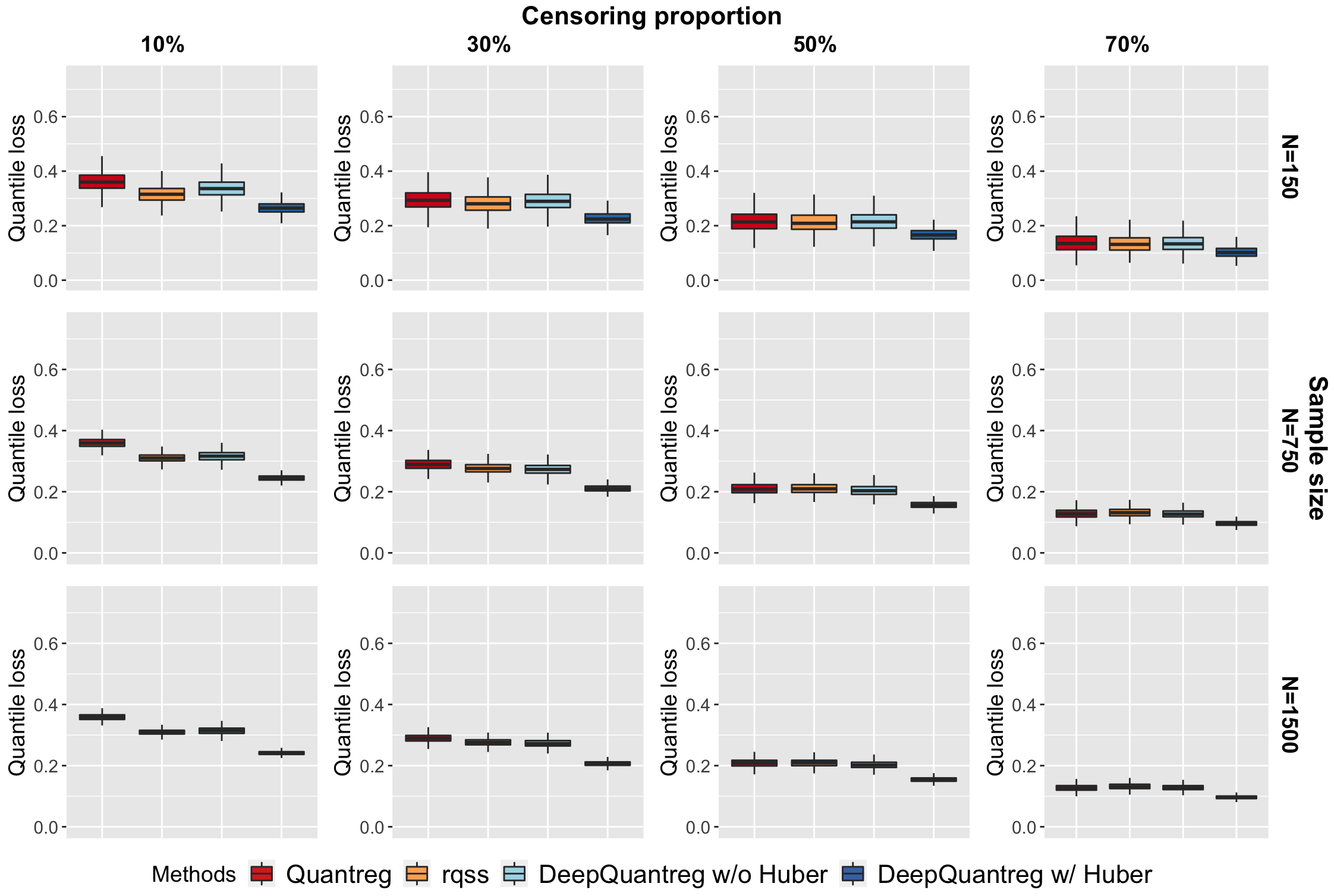}
  \caption{QL for group effect data}
  \label{fig:groupeffect_ql}
\end{subfigure}

\caption{Boxplot of $C$-Index, MMSE and quantile loss for different simulated data scenario when $\tau$ = 0.25}
\label{fig:boxplot25}
\end{figure}

\newpage

\begin{figure}
\begin{subfigure}{0.5\textwidth}
  \centering
  % include first image
  \includegraphics[width=.9\linewidth]{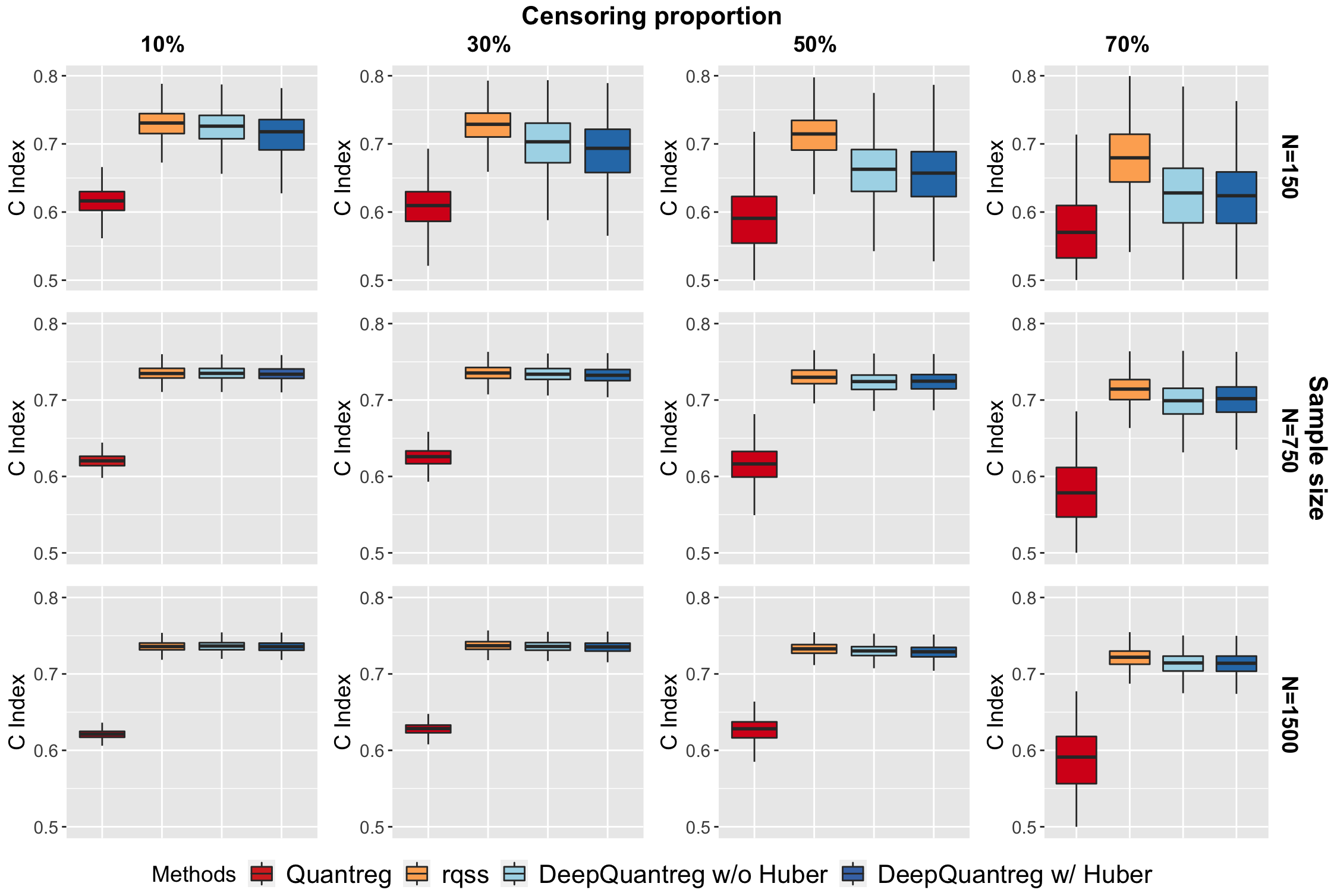}
  \caption{$C$-Index for no group effect data}
  \label{fig:noeffect_cindex}
\end{subfigure}
\begin{subfigure}{0.5\textwidth}
  \centering
  % include first image
  \includegraphics[width=.9\linewidth]{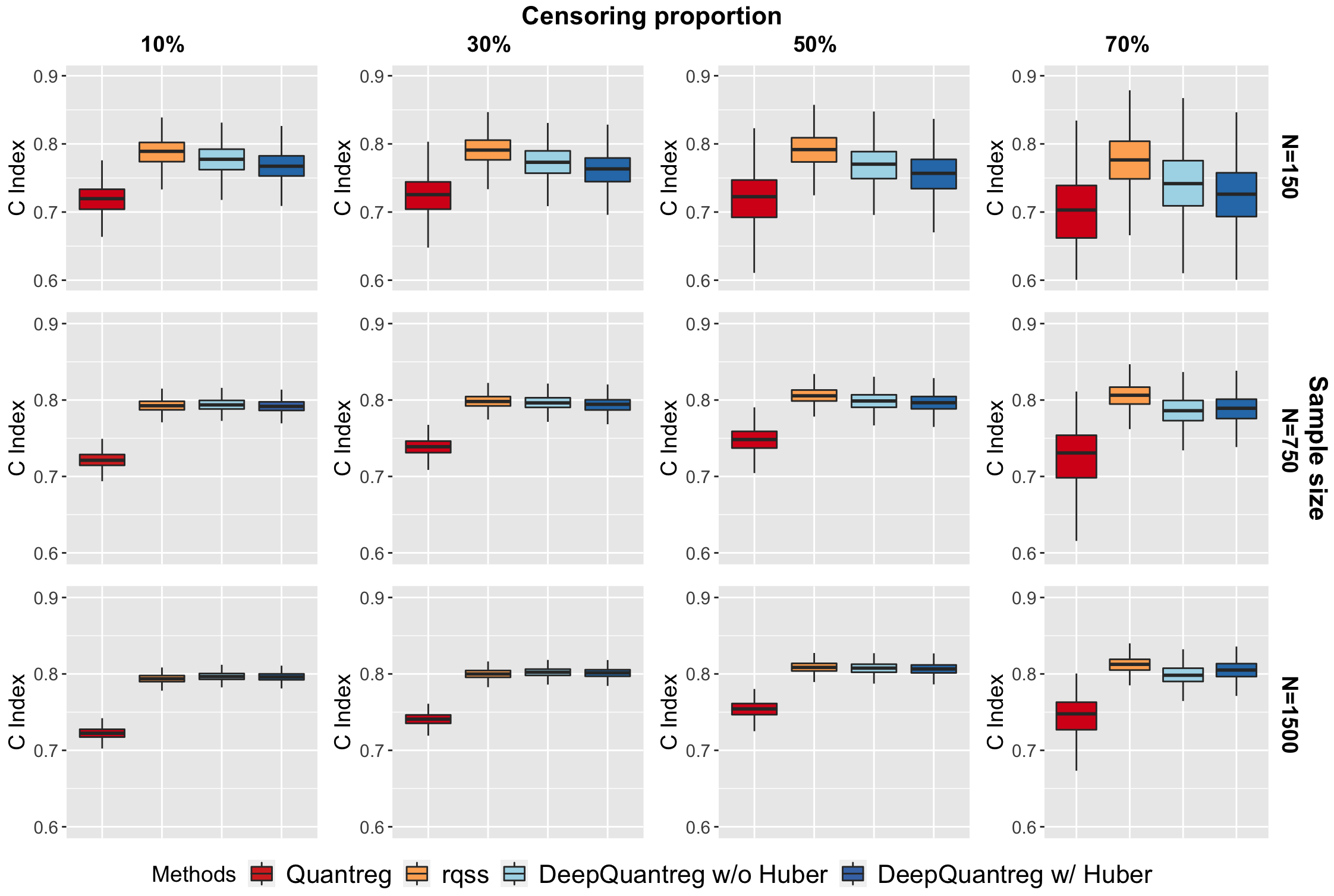}
  \caption{$C$-Index for group effect data}
  \label{fig:groupeffect_cindex}
\end{subfigure}

\begin{subfigure}{0.5\textwidth}
  \centering
  % include second image
  \includegraphics[width=.9\linewidth]{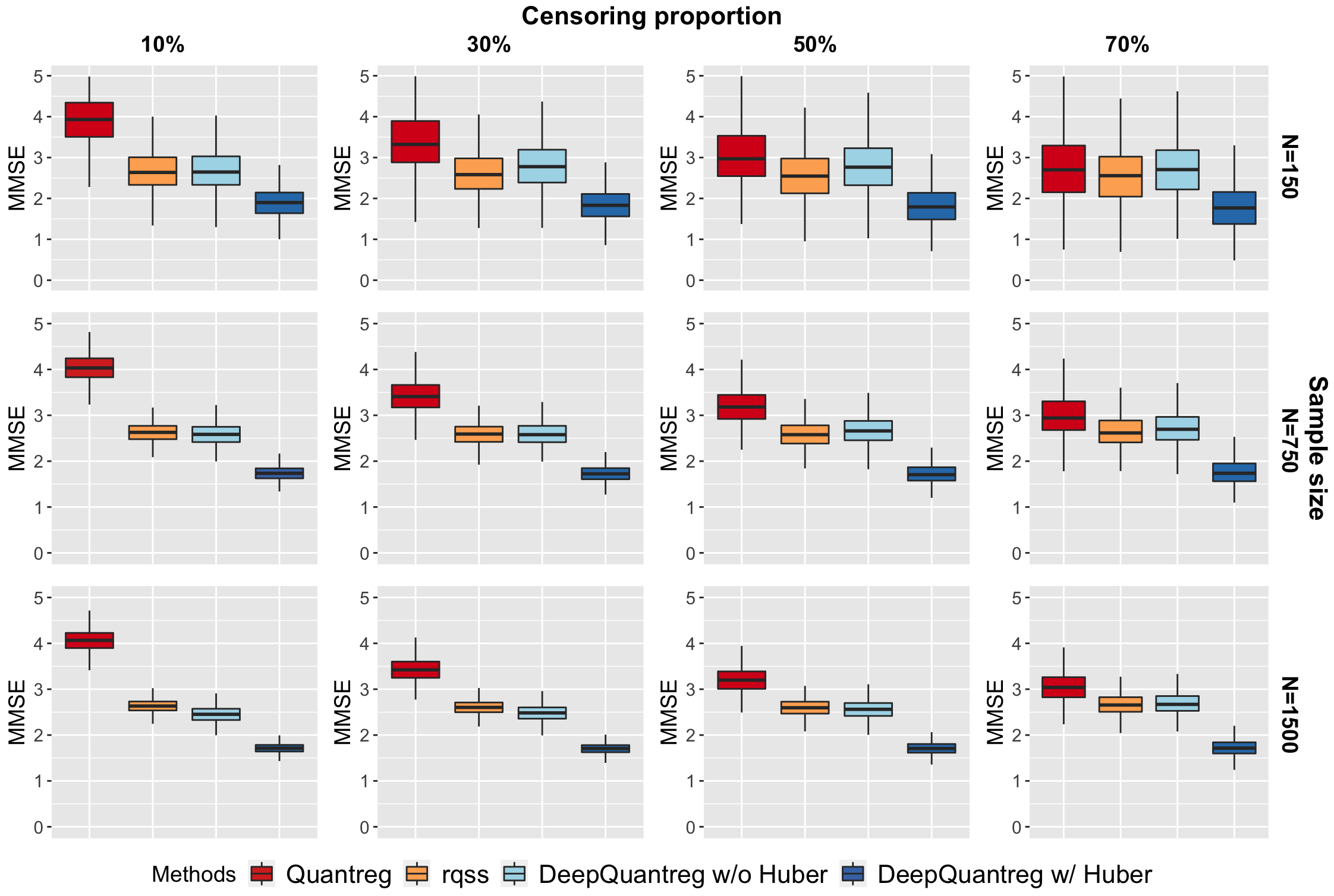}  
  \caption{MMSE for no group effect data}
  \label{fig:noeffect_mmse}
\end{subfigure}
\begin{subfigure}{0.5\textwidth}
  \centering
  % include first image
  \includegraphics[width=.9\linewidth]{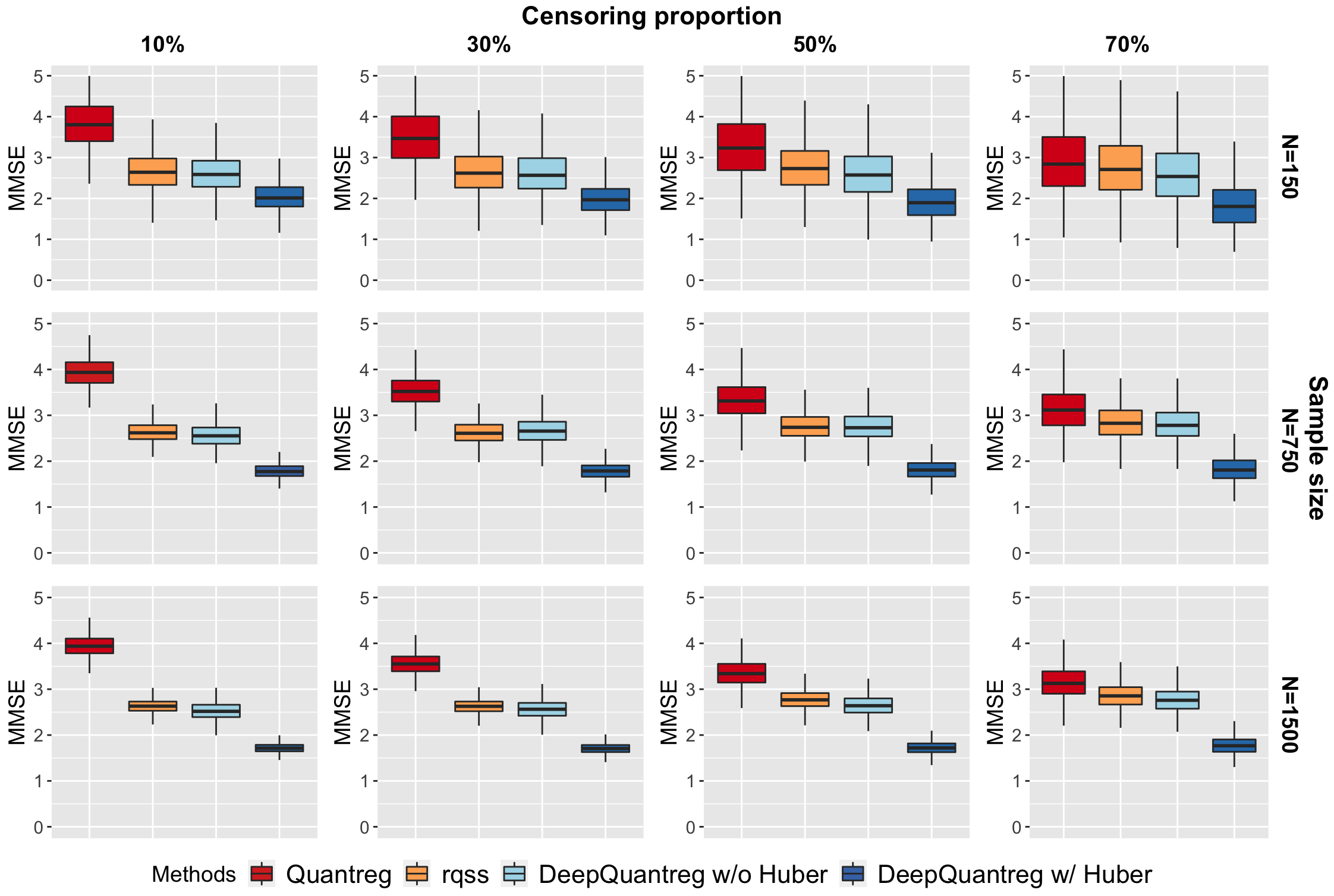}
  \caption{MMSE for group effect data}
  \label{fig:groupeffect_mmse}
\end{subfigure}

\begin{subfigure}{0.5\textwidth}
  \centering
  % include first image
  \includegraphics[width=.9\linewidth]{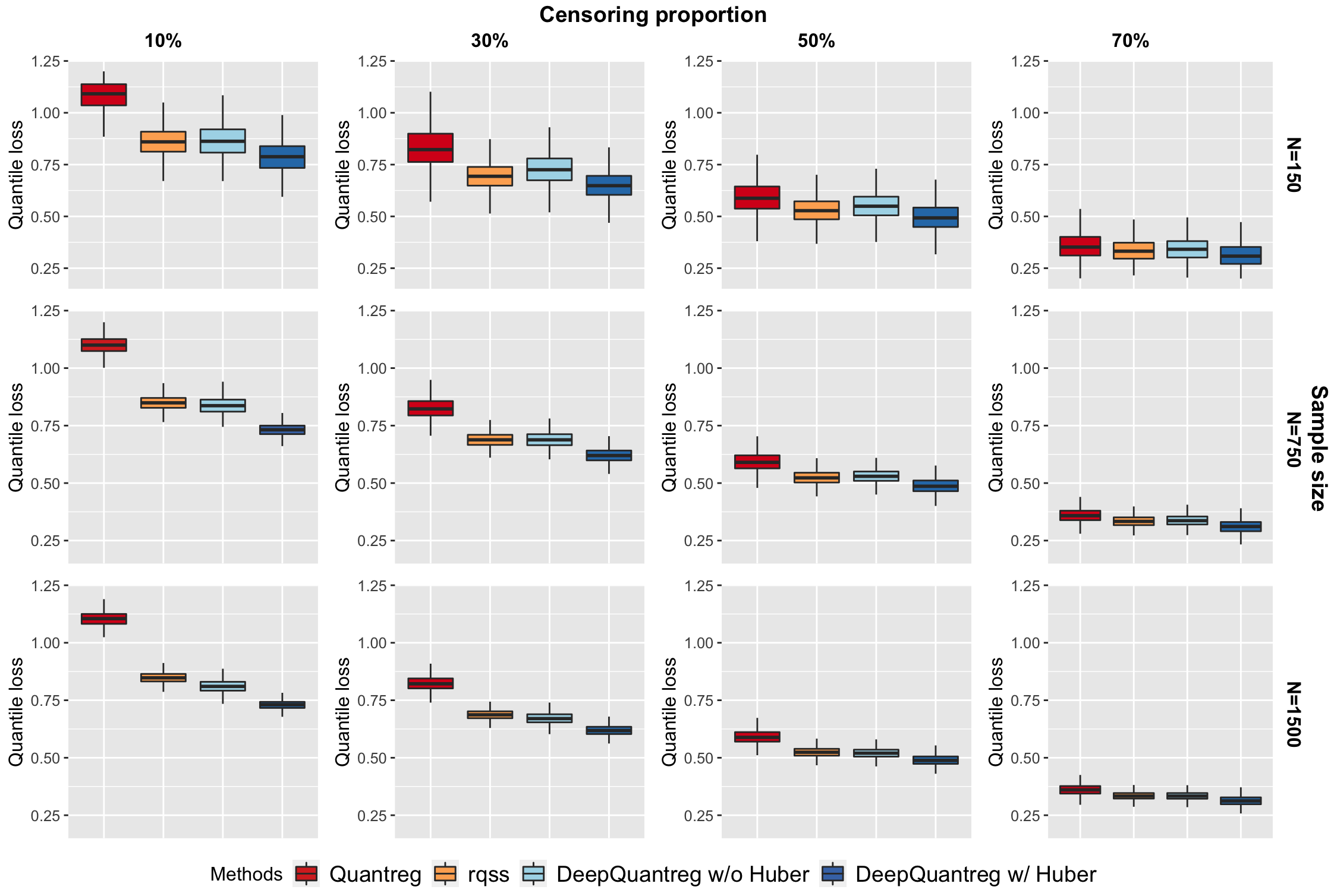}
  \caption{QL for no group effect data}
  \label{fig:noeffect_ql}
\end{subfigure}
\begin{subfigure}{0.5\textwidth}
  \centering
  % include first image
  \includegraphics[width=.9\linewidth]{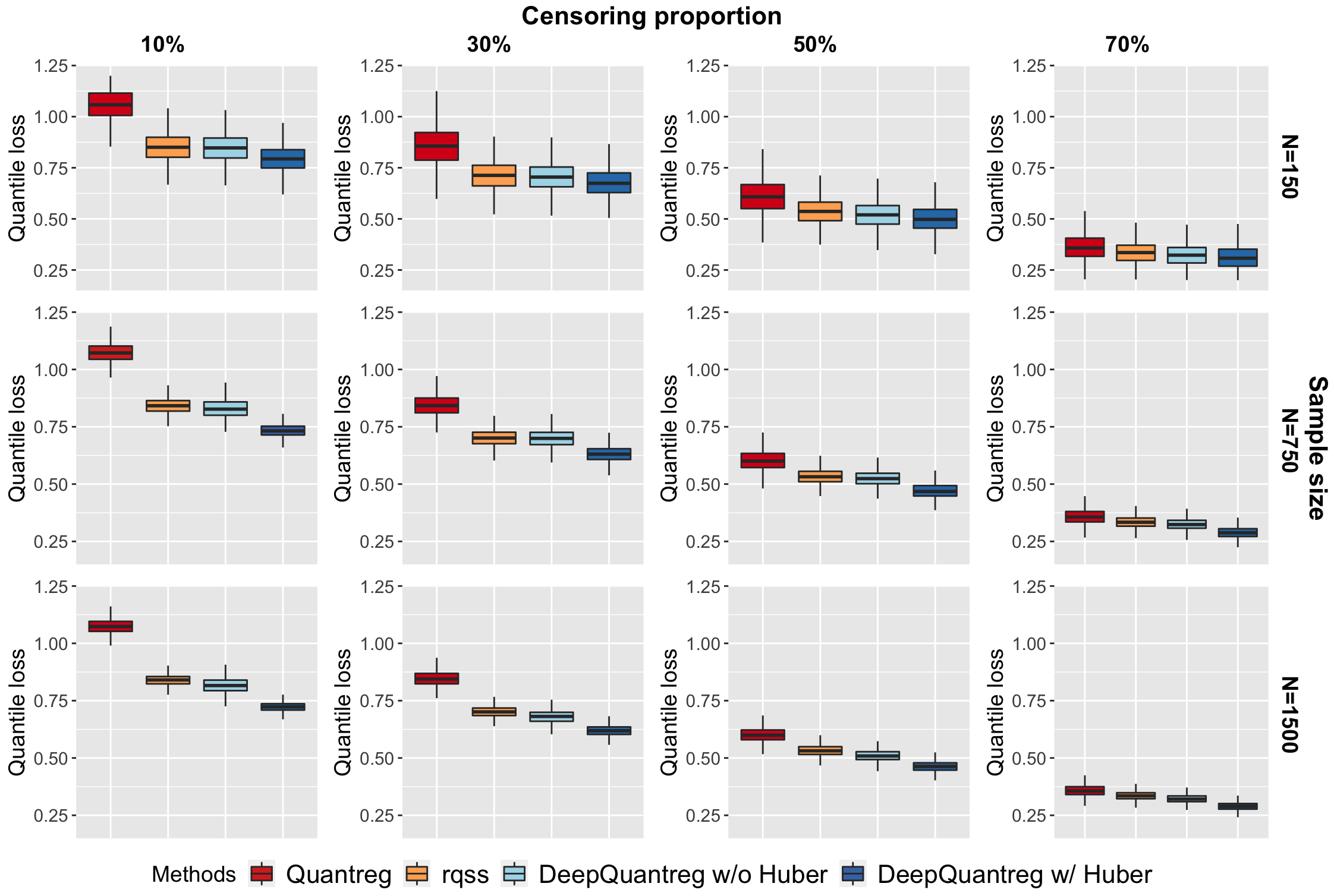}
  \caption{QL for group effect data}
  \label{fig:groupeffect_ql}
\end{subfigure}

\caption{Boxplot of $C$-Index, MMSE and quantile loss for different simulated data scenario when $\tau$ = 0.75}
\label{fig:boxplot75}
\end{figure}

\newpage

\begin{figure}
\begin{subfigure}{\textwidth}
  \centering
  % include first image
  \includegraphics[width=0.7\linewidth]{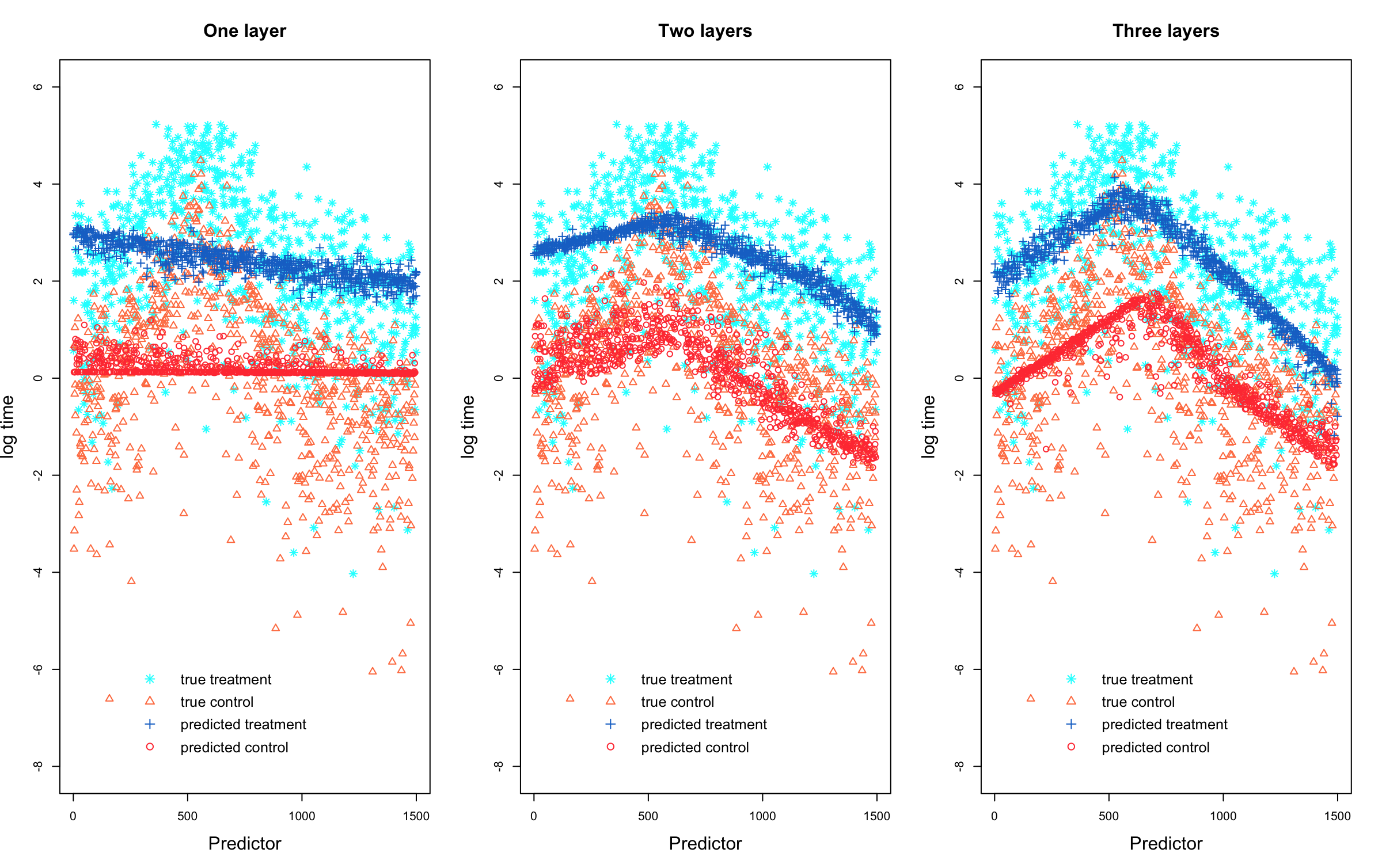}
  \caption{Prediction plots for different number of layers with 4 nodes/layer}
  \label{fig:4nodes}
\end{subfigure}

\begin{subfigure}{\textwidth}
  \centering
  % include third image
  \includegraphics[width=0.7\linewidth]{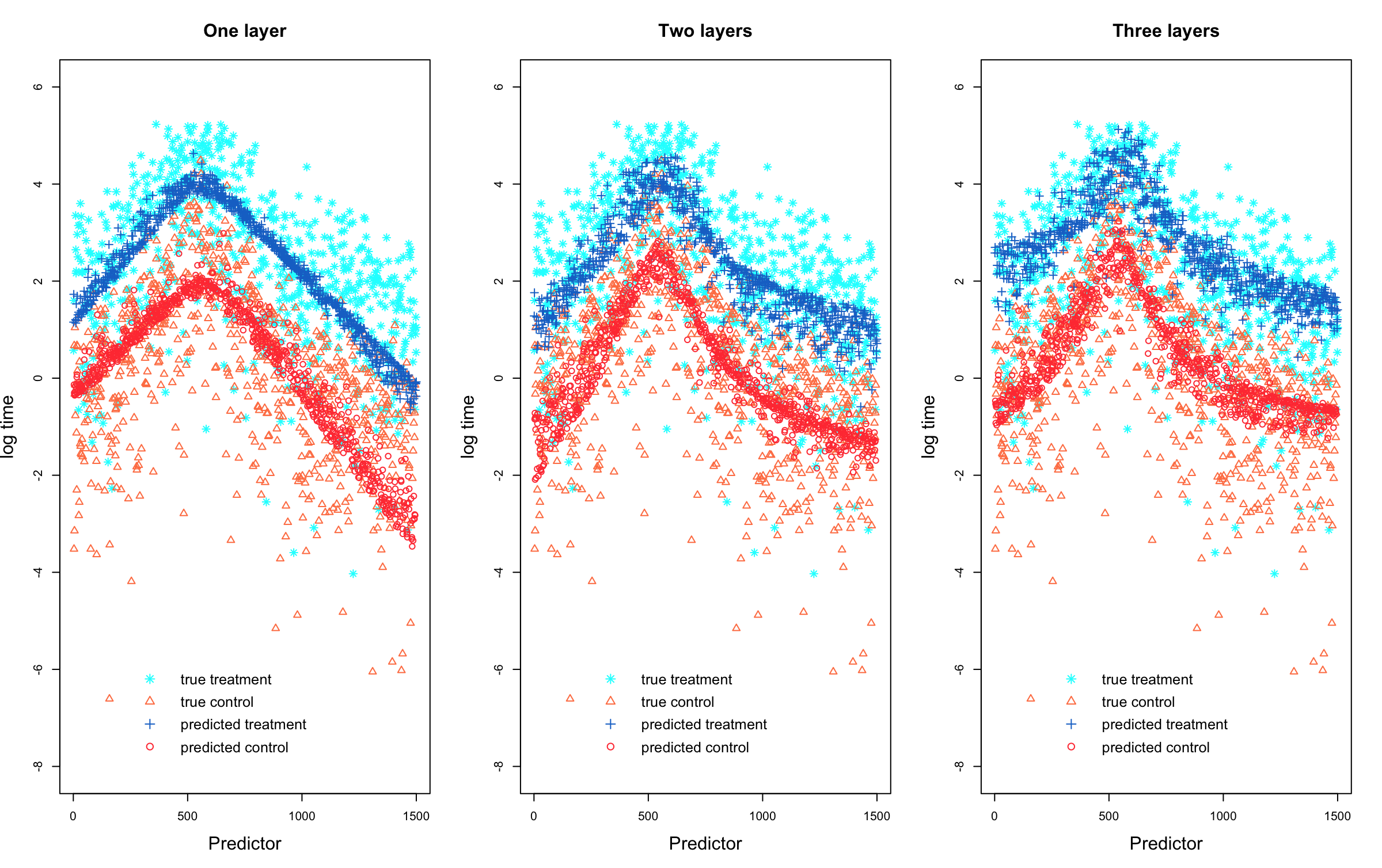}  
  \caption{Prediction plots for different number of layers with 300 nodes/layer}
  \label{fig:300nodes}
\end{subfigure}

\caption{Median ($\tau=0.5$) prediction plots for deep censored quantile regression model with different number of hidden layers and hidden nodes per layer using group effect data with 10\% censoring proportion}
\label{fig:Layercomparison}
\end{figure}

\newpage

\begin{figure}
    \centering
    \includegraphics[width=.8\linewidth]{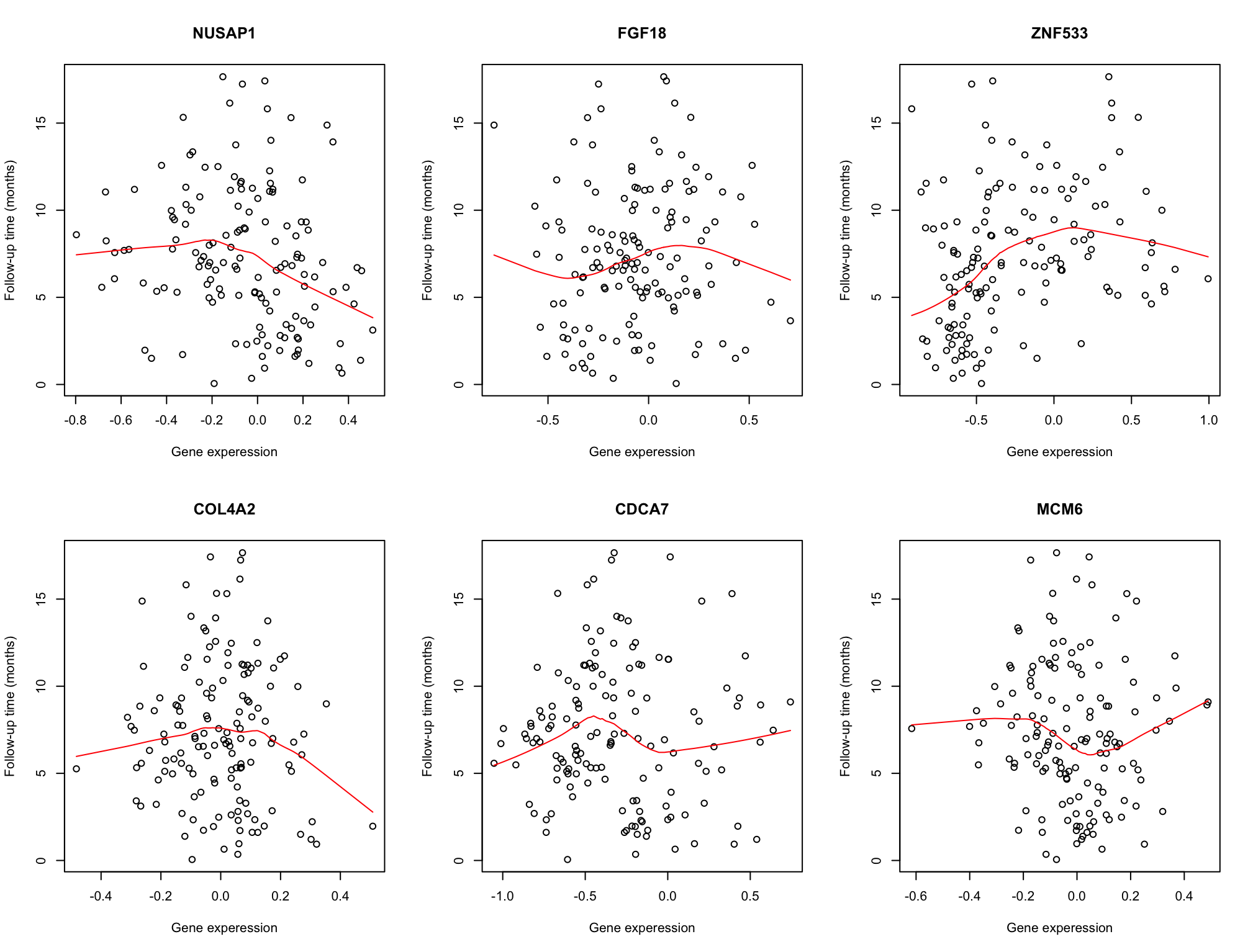}
    \caption{Scatter plots of gene expression with follow up time in NKI70 data}
    \label{fig:6genes}
\end{figure}

\clearpage

\begin{table}[h]
\centering
\caption{Hyperparameters used in simulation studies} 
\label{tab:hyperparameter}
\begin{threeparttable}
\begin{adjustbox}{width=1\textwidth}
\begin{tabular}{cccccccccc}
\hline
\multicolumn{2}{c}{} & \multicolumn{4}{c}{No effect data} & \multicolumn{4}{c}{Group effect data} \\
\multicolumn{2}{c}{censoring proportion} & 10\%           & 30\%                & 50\%     & 70\%         & 10\%              & 30\%         & 50\%         & 70\%    \\
\hline
\multicolumn{2}{c}{\# of layer} & \multicolumn{8}{c}{2} \\
\multicolumn{2}{c}{\# of nodes/layer}  & \multicolumn{8}{c}{300}\\
\multicolumn{2}{c}{Dropout rate}  & \multicolumn{4}{c}{0.3}& \multicolumn{4}{c}{0.4}\\
\multicolumn{2}{c}{Activation function}  & Sigmoid        & Hard sigmoid & Sigmoid &        Hard sigmoid & Hard sigmoid      & Hard sigmoid & Hard sigmoid & Sigmoid \\
\multicolumn{2}{c}{Optimizer}  & Nadam          & Adadelta            & Adadelta & Nadam        & Adam              & Adam         & Nadam        & Nadam   \\
\multirow{3}{*}{\# of epochs}   & n=150  & \multicolumn{4}{c}{500}& \multicolumn{4}{c}{1000}\\
& n=750  & \multicolumn{4}{c}{500}& \multicolumn{4}{c}{500} \\
& n=1500 & \multicolumn{4}{c}{500}& \multicolumn{4}{c}{500} \\
\multirow{3}{*}{Batch size}   & n=150  & \multicolumn{4}{c}{32}& \multicolumn{4}{c}{64}\\
& n=750  & \multicolumn{4}{c}{64}& \multicolumn{4}{c}{64} \\
& n=1500 & \multicolumn{4}{c}{64}& \multicolumn{4}{c}{64} \\
\hline
\end{tabular}
\end{adjustbox}
\begin{tablenotes}
      \tiny
      \item Acronyms: Adam - Adaptive Moment Estimation;

\hspace{1.2cm}  Nadam - Nesterov-accelerated Adaptive Moment Estimation
    \end{tablenotes}
\end{threeparttable}
\end{table}

\clearpage

\end{document}